%% file: main.tex
\documentclass[preprint,5p,times,twocolumn,authoryear]{elsarticle}

\usepackage{array}
\usepackage{makecell}
\usepackage{tabularx}
\usepackage{booktabs}
\usepackage{arydshln}
\usepackage[T1]{fontenc}
\usepackage[hyphens]{url}
\usepackage[hidelinks]{hyperref}
\usepackage[table]{xcolor}
\usepackage[nohyperlinks, nolist]{acronym}
\usepackage{amsmath}
\usepackage{amssymb}
\usepackage{calc}
\usepackage{csquotes}
\usepackage{pifont}
\usepackage{enumitem}
\usepackage[retain-explicit-plus]{siunitx}
\usepackage{subcaption}
\usepackage{wasysym}

\usepackage{latexml}
\iflatexml 
    \usepackage{authblk}
\fi

\DeclareMathOperator*{\argmax}{arg\,max}

\newcolumntype{M}[1]{>{\centering\arraybackslash}m{#1}}
\newcolumntype{L}[1]{>{\raggedright\arraybackslash}m{#1}}
\newcolumntype{R}[1]{>{\raggedleft\arraybackslash}m{#1}}

\newcommand*\affil[2][]{\address[#1]{#2}}

\newcommand{\cmark}{\ding{51}}

\IfFileExists{latexml.sty}{ 

\author[1]{Josafat-Mattias~Burmeister}

\author[2]{Andreas~Tockner}

\author[3]{Stefan~Reder}

\author[4]{Markus~Engel}

\author[1]{Rico~Richter}

\author[3]{Jan-Peter~Mund}

\author[1,5]{Jürgen~Döllner}

}{ 

\author[1]{Josafat-Mattias~Burmeister\corref{cor1}
\href{https://orcid.org/0000-0003-1890-844X}{\includegraphics[trim={0px 160px 0px 100px}, clip, height=0.95\baselineskip]{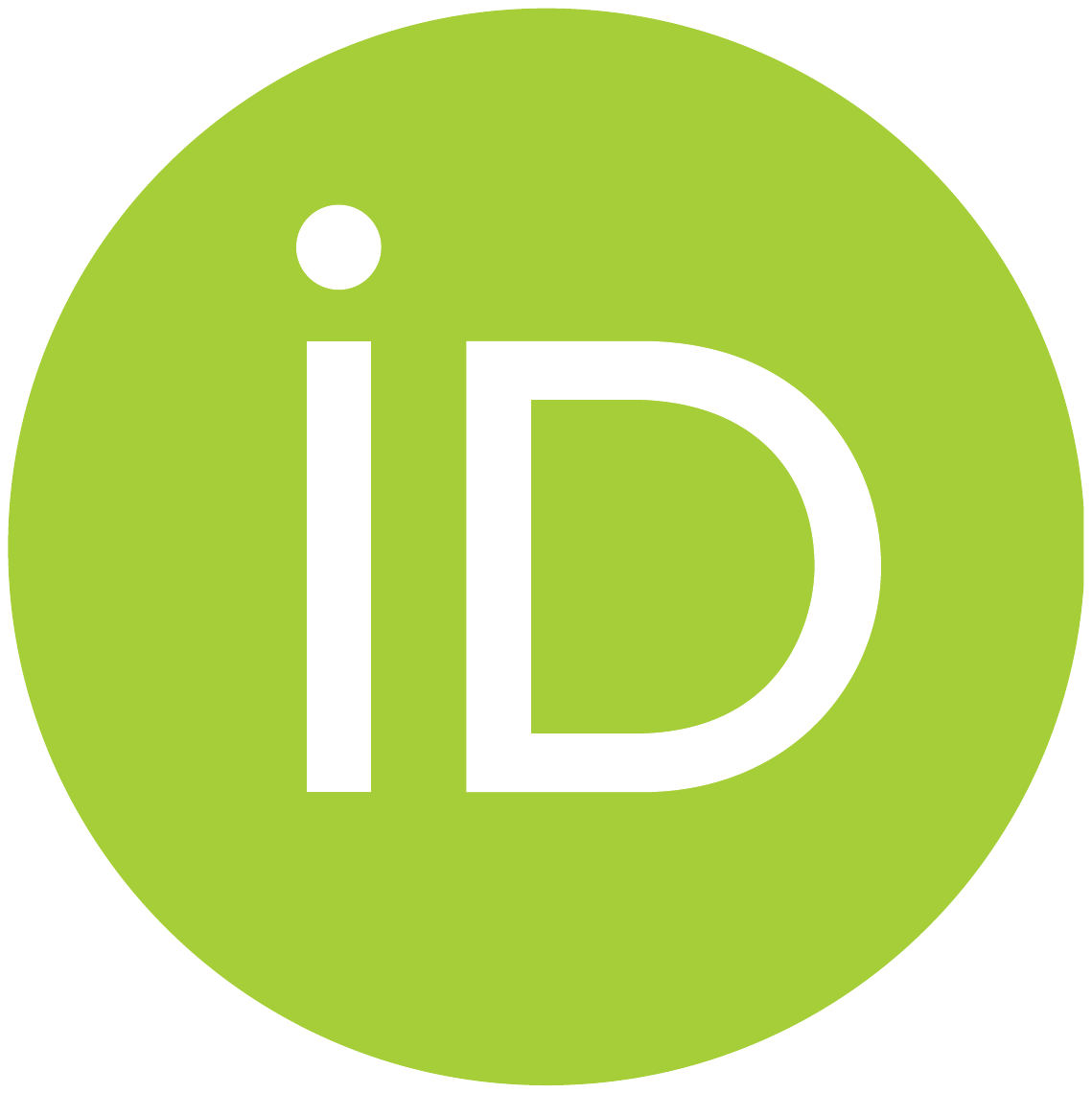}}
}

\author[2]{Andreas~Tockner
\href{https://orcid.org/0000-0001-6833-6713}{\includegraphics[trim={0px 160px 0px 100px}, clip, height=0.95\baselineskip]{img_ORCID_iD}}
}

\author[3]{Stefan~Reder
\href{https://orcid.org/0000-0002-2899-6191}{\includegraphics[trim={0px 160px 0px 100px}, clip, height=0.95\baselineskip]{img_ORCID_iD}}
}

\author[4]{Markus~Engel
\href{https://orcid.org/0000-0001-6991-9021}{\includegraphics[trim={0px 160px 0px 100px}, clip, height=0.95\baselineskip]{img_ORCID_iD}}
}

\author[1]{Rico~Richter
\href{https://orcid.org/0000-0001-5523-3694}{\includegraphics[trim={0px 160px 0px 100px}, clip, height=0.95\baselineskip]{img_ORCID_iD}}
}

\author[3]{Jan-Peter~Mund~\href{https://orcid.org/0000-0002-4878-5519}{\includegraphics[trim={0px 160px 0px 100px}, clip, height=0.95\baselineskip]{img_ORCID_iD}}
}

\author[1,5]{Jürgen~Döllner
\href{https://orcid.org/0000-0002-8981-8583}{\includegraphics[trim={0px 160px 0px 100px}, clip, height=0.95\baselineskip]{img_ORCID_iD}}
}

\cortext[cor1]{Corresponding author}

}

\affil[1]{
    University of Potsdam, Digital Engineering Faculty, Prof.-Dr.-Helmert-Straße~2-3, Potsdam, 14482, Germany
}

\affil[2]{
    {BOKU} University, Department of Ecosystem Management, Climate and Biodiversity, Institute of Forest Growth, Peter-Jordan-Straße~82, Wien, 1190, Austria
}

\affil[3]{
    Eberswalde University of Sustainable Development, Department of Forestry, Alfred-Möller-Straße~1, Eberswalde, 16225, Germany
}

\affil[4]{
    State Forestry Research Centre, Alfred-Möller-Straße~1, Eberswalde, 16225, Germany
}

\affil[5]{
    Hasso Plattner Institute for Digital Engineering, Prof.-Dr.-Helmert-Straße~2-3, Potsdam, 14482, Germany
}

\acrodef{ALS}{airborne laser scanning}
\acrodef{CCI}{circumferential completeness index}
\acrodef{CSF}{cloth simulation filtering}
\acrodef{DBH}{diameter at breast height}
\acrodef{DTM}{digital terrain model}
\acrodef{FN}{false negative}
\acrodef{FP}{false positive}
\acrodef{GAM}{generalized additive model}
\acrodef{IoU}{Intersection over Union}
\acrodef{mIoU}{mean Intersection over Union}
\acrodef{MLS}{mobile laser scanning}
\acrodef{mPrecision}{mean precision}
\acrodef{mRecall}{mean recall}
\acrodef{PLS}{personal laser scanning}
\acrodef{RANSAC}{Random Sample Concensus}
\acrodef{TLS}{terrestrial laser scanning}
\acrodef{TP}{true positive}
\acrodef{ULS}{UAV-borne laser scanning}

\begin{document}

\onecolumn
\section*{Graphical Abstract - treeX: Unsupervised Tree Instance Segmentation in Dense Forest Point Clouds}
\input{sec_0_graphical_abstract}
\twocolumn

\begin{frontmatter}

\title{treeX: Unsupervised Tree Instance Segmentation in Dense Forest Point Clouds} 

\begin{abstract}
\input{sec_0_abstract}
\end{abstract}

\begin{keyword}
3D Point Clouds, Tree Instance Segmentation, LiDAR, Forest, Python Package
\end{keyword}

\end{frontmatter}

\section{Introduction}
\label{sec:introduction}
\input{sec_1_introduction}

\section{Related Work}
\label{sec:related_work}
\input{sec_2_related_work}

\section{{treeX} Algorithm}
\label{sec:algorithm}
\input{sec_3_algorithm}

\section{Evaluation Methods}
\label{sec:evaluation}
\input{sec_4_evaluation}

\section{Results}
\label{sec:results}
\input{sec_5_results}

\section{Discussion}
\label{sec:discussion}
\input{sec_6_discussion}

\section{Conclusions}
\label{sec:conclusions}
\input{sec_7_conclusions}

\section*{Declaration of Generative AI and AI-Assisted Technologies in the Writing Process}
\label{sec:ai-disclosure}
\input{sec_8_ai_disclosure}

\section*{CRediT Authorship Contribution Statement}
\label{sec:authorship}
\input{sec_9_authorship}

\section*{Declaration of Competing Interest}
\label{sec:competing-interest}
\input{sec_10_declaration_of_competing_interest}

\section*{Acknowledgements}
\label{sec:acknowledgements}
\input{sec_11_acknowledgements}

\section*{Data and Code Availability}
\label{sec:data-availability}
\input{sec_12_data_availability}

\appendix
\input{sec_13_appendix}

\bibliographystyle{elsarticle-harv} 
\bibliography{main.bib}

\end{document}

%% file: sec_0_graphical_abstract.tex
\begin{figure*}[!h]
    \centering
    \includegraphics{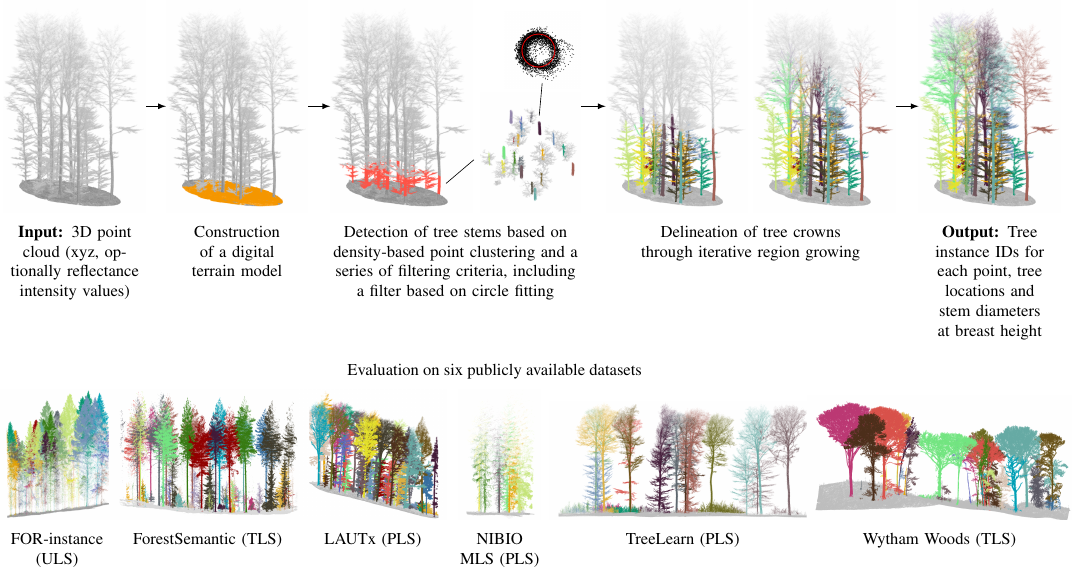}
\end{figure*}
\vfill

%% file: sec_0_abstract.tex
Close-range laser scanning provides detailed 3D~captures of forest stands but requires efficient software for processing 3D~point cloud data and extracting individual trees.
Although recent studies have introduced deep learning methods for tree instance segmentation, these approaches require large annotated datasets and substantial computational resources.
As a resource-efficient alternative, we present a revised version of the treeX algorithm, an unsupervised method that combines clustering-based stem detection with region growing for crown delineation.
While the original treeX algorithm was developed for personal laser scanning (PLS) data, we provide two parameter presets, one for ground-based laser scanning (stationary terrestrial - TLS and PLS), and one for UAV-borne laser scanning (ULS).
We evaluated the method on six public datasets (FOR-instance, ForestSemantic, LAUTx, NIBIO MLS, TreeLearn, Wytham Woods) and compared it to six open-source methods (original treeX, treeiso, RayCloudTools, ForAINet, SegmentAnyTree, TreeLearn).
Compared to the original treeX algorithm, our revision reduces runtime and improves accuracy, with instance detection F$_1$-score gains of +\num{0.11} to +\num{0.49} for ground-based data.
For ULS data, our preset achieves an F$_1$-score of \num{0.58}, whereas the original algorithm fails to segment any correct instances.
For TLS and PLS data, our algorithm achieves accuracy similar to recent open-source methods, including deep learning.
Given its algorithmic design, we see two main applications for our method: (1) as a resource-efficient alternative to deep learning approaches in scenarios where the data characteristics align with the method design (sufficient stem visibility and point density), and (2) for the semi-automatic generation of labels for deep learning models.
To enable broader adoption, we provide an open-source Python implementation in the pointtree package.

%% file: sec_1_introduction.tex
The measurement of forest stands with close-range laser scanning provides detailed 3D~point clouds, from which tree- and stand-level parameters can be deduced \citep{fassnacht-2024,calders-2020}. The information that can be obtained from 3D~point clouds encloses stem diameters and tree heights \citep{gollob-2020}, crown architectures \citep{knapp-wilson-2023}, as well as estimates of above-ground biomass \citep{borsah-2023} and radiative transfer \citep{witzmann-2024}. There are a wide variety of potential applications in the field of forest management, and 3D~point cloud analysis is starting to be applied to large-scale forest inventories \citep{persson-2022} and monitoring \citep{goodbody-2021}. 
Measuring tree structural components from laser scanning data of forest ecosystems requires a segmentation of unlabeled 3D~data, which can be discriminated into two tasks: the segmentation of individual trees, referred to as instance segmentation \citep{latifi-2015}, and the classification of their structural components (e.g., stems, branches, and foliage) and non-tree elements (e.g., ground, understory, and man-made objects), referred to as semantic segmentation \citep{krisanski-2021a}.
The design of segmentation methods can vary depending on the type of sensor and the platform used to capture the 3D~point cloud data, which define the quality, point density, and per-point attributes (e.g., reflectance intensity).
Ground-based scanning of forests with stationary devices (terrestrial laser scanning - TLS\acused{TLS}) or portable handheld devices (personal laser scanning - PLS\acused{PLS}) provides accurate representations of tree stems and lower parts of the forest canopy, while providing less accurate information on the upper tree crown \citep{tockner-2022}. In contrast, scanning with airborne sensors (e.g., UAV-borne laser scanning - ULS\acused{ULS}) above the forest canopy provides accurate information of the upper tree crown, but less detailed information on tree stems and forest understory \citep{nemmaoui-2024}. While sparse airborne 3D~point clouds may not provide sufficient information for tree instance and semantic segmentation, dense 3D~point clouds (with a point density usually far beyond \num{1000}~pts/m$^2$) can be obtained by merging scans from various sensors, such as personal and airborne laser scanning devices \citep{panagiotidis-2022,buelbuel-2025}.

Tree instance segmentation of 3D point clouds can be achieved using deep learning \citep{xiang-2024,henrich-2024,wielgosz-2024,xi-2025} or unsupervised algorithmic approaches \citep{bienert-2021,lowe-2021,xi-2022,tockner-2022}. Deep learning approaches attempt to learn patterns from manually segmented training data using various model architectures \citep{guo-2021}, such as convolutional neural networks \citep{wielgosz-2024} or transformer-based networks \citep{xi-2025}. Although deep learning approaches can be very accurate, their accuracy depends on the availability of sufficient training data with labeled instances, and the patterns learned by deep learning models are difficult to interpret due to their black-box nature \citep{mulawade-2024}. Furthermore, training and inference of deep learning models are often reliant on powerful computation environments.
Algorithmic approaches attempt to segment tree instances through the rule-based combination of shape detection methods and statistical tools \citep{bienert-2021,burt-2019}, such as iterative region growing \citep{tockner-2022}, or using graphs constructed from 3D~point clouds \citep{wang-2021,wilkes-2023,xu-2023}. These algorithmic approaches are applicable for semi-automatic segmentation without relying on training data first-hand. Their rule-based architecture facilitates their interpretation and evaluation, with a certain degree of customizability. Due to these properties, they are a valuable alternative to deep learning methods. Furthermore, unsupervised approaches might also support the development of deep learning methods by generating pseudo-labels for model pre-training~\citep{henrich-2024}. In particular, compared to self-training approaches, which rely on a deep learning model’s own predictions, pseudo-labels derived from unsupervised methods may help mitigate pre-training bias~\citep{chen-2022}.

However, there are relatively few unsupervised tree instance algorithms with publicly available, computationally efficient, and well-tested implementations that can be readily applied to custom datasets (Table~\ref{tab:related-work}).
Therefore, this study presents an unsupervised algorithm for segmenting tree instances in dense 3D~point clouds. The presented method is a revised version of the treeX algorithm originally introduced by \citet{tockner-2022} and \citet{gollob-2020}. While the original treeX algorithm was developed with a focus on \ac{PLS} data, we provide two different parameter presets, one for \ac{TLS} and \ac{PLS} data and one for \ac{ULS} data, to support broader applications and various sensors. The method offers a good balance between segmentation accuracy and computational efficiency and is implemented in a well-tested and well-documented open-source Python package. It was evaluated using various data from six publicly available datasets. Although the method's generalizability is limited to scenarios with sufficient stem visibility and point density, in such conditions, it proves to be a valuable alternative to deep learning approaches.

%% file: sec_2_related_work.tex
In the following, we summarize existing methods for segmenting individual trees in forest point clouds, which form the context of our study and serve as references in our evaluation.
These methods can be broadly categorized into deep learning approaches and algorithmic methods (Table~\ref{tab:related-work}), with recent research focusing predominantly on the former.

\subsection{Deep Learning Tree Instance Segmentation Approaches}

Deep learning tree instance segmentation approaches can be further categorized into 2D and 3D~methods. In 2D~methods, 3D~point clouds are converted into a 2D raster representation (e.g., a canopy height model) and processed using 2D~deep learning architectures from the broad field of image classification. In contrast, 3D~methods operate directly on the point cloud data or on voxel-based 3D~representations derived from it, which makes them computationally more expensive.

\paragraph{2D~Methods} Among these methods are the works of \citet{chang-2022} and \citet{straker-2023}. \citet{chang-2022} propose a two-step deep learning approach that uses a 3D~deep learning method for semantic segmentation and a 2D~method for tree instance segmentation in \ac{TLS} forest scans. Semantic segmentation is achieved using the RandLA-Net architecture~\citep{hu-2020}, which classifies points as either tree or non-tree points. Tree instance segmentation is accomplished using 2D~feature maps calculated from 3D~points. The feature maps consist of three normalized channels, which characterize each grid cell by its maximum z-coordinate, the difference between its maximum and minimum z-coordinate, and the number of points contained within it. The YOLOv3 architecture~\citep{redmon-2016} is used to predict the bounding boxes of individual trees. Subsequently, the 2D~results are projected onto the 3D~point cloud and refined using 3D~clustering algorithms.

\citet{straker-2023} introduce a similar 2D~method for \ac{ULS} data. Their approach involves constructing a normalized canopy height model from the 3D point cloud and transforming it into a pseudo-color RGB image. The YOLOv5 architecture~\citep{redmon-2016} is then used to predict the bounding boxes of individual trees, which are projected onto the 3D~point cloud.

\input{tab_0_related_work}

\paragraph{3D~Methods} Several existing 3D~deep learning methods for tree instance segmentation are based on the PointGroup architecture~\citep{jiang-2020}, which implements a bottom-up instance segmentation approach that predicts offset and embedding vectors for each point and uses them to cluster the points into individual instances. Offset vectors are trained leading from each point toward the center of its respective instance. Embedding vectors are learned so that points belonging to the same instance lie close together in the embedding space and are distant from points belonging to other instances.
Tree instance segmentation approaches based on the PointGroup architecture include ForAINet~\citep{xiang-2024}, SegmentAnyTree~\citep{wielgosz-2024}, and TreeLearn~\citep{henrich-2024}.

ForAINet \citep{xiang-2024} is a deep learning model trained on the FOR-instance dataset~\citep{puliti-2023} that aims to generalize across \ac{ULS} point clouds from various forest types. The model is a 3D~convolutional neural network with two instance segmentation heads that predict offset and embedding vectors for each point. Two semantic segmentation heads segment points coarsely into tree and non-tree points and further into ground, low vegetation, stem, dead branches, and living branches. Additionally, a ScoreNet module predicts a quality score for each instance proposal. Experimenting with different hyperparameter settings and data augmentation techniques revealed the benefits of synthesizing training point clouds by mixing trees from various datasets. With this approach, ForAINet outperforms treeiso \citep{xi-2022}, Point2Tree \citep{wielgosz-2023}, and the 2D method proposed by \cite{straker-2023}.

SegmentAnyTree adopts the ForAINet architecture and trains it on a larger dataset to provide a model applicable to various types of sensors and platforms. The FOR-instance \citep{puliti-2023} and NIBIO MLS \citep{wielgosz-2023} datasets are used for training, testing different levels of point sparsification. Model evaluation was performed on the datasets Wytham Woods \citep{calders-2022}, TreeLearn \citep{henrich-2024}, LAUTx \citep{tockner-2022}, NIBIO MLS \citep{wielgosz-2023}, and FOR-instance \citep{puliti-2023}, where Segment\-Any\-Tree was able to outperform other approaches on most datasets.
The results indicate that random subsampling is an effective data augmentation strategy, as it enabled the model to maintain consistent performance across different point cloud resolutions. A noticeable decline in performance was observed only at very low point cloud densities (\(<\)~\num{50}~pts/m$^2$).

TreeLearn \citep{henrich-2024} uses a simplified version of the PointGroup architecture, incorporating only an offset prediction head for instance segmentation and a binary semantic segmentation head that classifies points as either tree or non-tree. The TreeLearn model was pre-trained on \ac{PLS} data automatically labeled using the software LiDAR360. Segmentation performance was greatly improved by further model training with manually segmented tree data from different sensors. The approach outperformed SegmentAnyTree \citep{wielgosz-2024}, ForAINet \citep{xiang-2024}, and TLS2trees \citep{wilkes-2023} on the TreeLearn dataset, a \ac{PLS} dataset newly introduced by the authors, and the Wytham Woods dataset \citep{calders-2022}.

\citet{xiang-2025} further develop PointGroup-based approaches by introducing ForestFormer3D, a model that performs tree instance segmentation and semantic segmentation, classifying points as ground, wood, or leaf. The model employs a query decoder that applies cross-attention between point-cloud embeddings and instance or semantic queries. Instance queries are generated using an offset prediction and a binary semantic segmentation head, similar to those used in the PointGroup architecture. The model is trained using \ac{TLS}, \ac{PLS}, and \ac{ULS} data and outperforms TreeLearn and ForAINet on FOR-instanceV2, an extended benchmark dataset newly introduced by the authors.

In contrast to the aforementioned end-to-end learning approaches, \cite{xi-2025} introduce a four-stage deep learning pipeline called TreeisoNet: In the first stage, stem and non-stem points are segmented using a SegFormer model. The second stage employs another SegFormer model to predict stem base locations, represented as a binary 2D~grid. In the third stage, a convolutional neural network predicts 2D~offset vectors from each stem point toward the corresponding stem base, which are then used to cluster stem points into individual instances. Finally, 3D~offset vectors are predicted for each non-stem point, shifting them toward the nearest stem point of the corresponding tree instance. Model training and evaluation were conducted separately for \ac{TLS}, \ac{ULS}, and \ac{ALS} data, with the TreeisoNet pipeline slightly outperforming ForAINet on the \ac{ULS} data and slightly underperforming treeiso~\citep{xi-2022} on the \ac{TLS} data. Substantial performance improvements were observed when the intermediate results from each pipeline stage were manually corrected.

\citet{xiu-2025} propose 3DPS-Net, a tree instance segmentation network that can be prompted to segment tree instances at a given position in 3D space. The network combines a KPConv-based~\citep{thomas-2019} point-cloud encoder and decoder with a prompt encoding module. The model was trained using \ac{TLS} and \ac{PLS} data. To generate prompts, both interactive prompting, which uses tree positions as prompts, and random generation of prompt points were tested.

\subsection{Algorithmic Tree Instance Segmentation Approaches}

Algorithmic tree instance segmentation approaches can be broadly categorized into graph-based methods and methods based on clustering and region growing.

\paragraph{Graph-Based Methods} In graph-based methods, point clouds are transformed into graph representations, where individual points or clusters of points constitute the graph nodes, and edges reflect spatial adjacency. Typically, the stem segments are identified first, and the remaining nodes are assigned to the corresponding stems using connectivity criteria and shortest path algorithms.

For example, \cite{wang-2021} introduce a graph-pathing method to extract trees from \ac{TLS} point clouds. They construct a hybrid graph from individual points using a combination of k-nearest neighbors and Delaunay triangulation. The algorithm identifies the locally lowest nodes in this graph as tree base candidates (root nodes). Then, from each node, it finds a path to one of these root nodes using downward traversal and shortest-path criteria. All nodes that reach the same root node are assigned to the same tree. This method has been tested on \ac{TLS} datasets from Finland \citep{liang-2024} and Australia \citep{calders-2022}.

RayCloudTools~\citep{lowe-2021}, a toolset for 3D mapping, visualization, and analysis of ray clouds, includes a similar graph-based algorithm. The algorithm computes the shortest path from each point to the ground. The endpoints of these paths are binned into a 2D grid, and points whose paths terminate in the same or adjacent grid cells are grouped as belonging to the same tree. The underlying graph structure is then used to reconstruct a tree model, where individual branch sections are represented as cylinders.

TLS2trees~\citep{wilkes-2023} provides a graph-based tree instance segmentation algorithm that builds upon a deep learning-based semantic segmentation into ground, wood, leaf, and coarse woody debris~\citep{krisanski-2021a}. A graph is constructed from wood points, where nodes represent point clusters extracted from different horizontal layers. Root nodes corresponding to tree stems are identified using RANSAC cylinder fitting. From each node, the shortest path to a root node is computed, and nodes sharing the same root are grouped into tree instances. A second graph is then constructed from the leaf points, and each leaf point is assigned to a tree instance based on the shortest path to a branch tip. The algorithm was evaluated using ten \ac{TLS} scans covering different forest types. \citet{wielgosz-2023} present a framework called Point2Tree to optimize the hyperparameters of the method, improving overall performance by four percentage points.

Treeiso \citep{xi-2022} is a graph-cut algorithm for segmenting individual trees from \ac{TLS} point clouds. The algorithm constructs a graph using the k-nearest neighbor method and initially over-segments it into small point clusters using the Cut-Pursuit graph-cut algorithm in 3D. By applying the Cut-Pursuit algorithm in 2D, these clusters are subsequently merged into larger segments. The final tree instances are then formed by grouping these segments based on adjacency and connectivity criteria.

\citet{xu-2023} propose a topology-based tree segmentation algorithm based on discrete Morse theory. The method constructs a graph from the point cloud and initially over-segments it into a collection of minimum ascending regions. Regions corresponding to tree bases are then identified and expanded into full tree instances through graph analysis. The algorithm was evaluated on several \ac{TLS} datasets, including Wytham Woods~\citep{calders-2022}, and demonstrated superior performance compared to 3D Forest~\citep{trochta-2017} and FSCT~\citep{krisanski-2021b}.

\paragraph{Methods Based on Clustering and Region Growing} Several tree instance segmentation algorithms rely on bottom-up clustering or region growing techniques. Typically, these approaches initially cluster points within the stem layer to separate individual tree stems. Subsequently, the stem segments are iteratively expanded through additional clustering or region growing operations to delineate the corresponding tree crowns.

For example, the 3D~Forest software \citep{trochta-2017} includes a clustering approach for tree instance segmentation in \ac{TLS} point clouds. The algorithm divides the point cloud into multiple horizontal layers and clusters the points within each slice using Euclidean clustering. Clusters of different layers are merged on the basis of a distance criterion to obtain the individual stems. The stem segments are iteratively expanded by adding remaining clusters based on distance criteria and growth direction. The algorithm was evaluated using a \ac{TLS} dataset.

Similarly, \citet{fu-2022} introduce a clustering method that uses the DBSCAN algorithm to detect tree stem candidates and filters them using Hough circle fitting. To determine suitable clustering parameters, an approach based on a distance distribution matrix is proposed. To delineate tree crowns, fuzzy K-means clustering is performed on different point cloud layers, and the initial cluster positions are initialized based on the detected stem locations. The algorithm was evaluated using two \ac{TLS} datasets.

Also targeting \ac{TLS} data, treeseg \citep{burt-2019} uses a combination of Euclidean clustering, normal-based region growing, RANSAC cylinder fitting, and verticality analysis to detect tree stems. To segment tree crowns, cylinders are placed on the identified stem locations, and the points within these cylinders are segmented based on connectivity testing.

\citet{bienert-2021} propose a tree instance segmentation algorithm for \ac{MLS} data. The algorithm detects tree stems using voxel-based clustering, followed by a filtering of the clusters based on circle fitting. To delineate individual tree crowns, the detected stem clusters are iteratively expanded using a region growing approach. Remaining crown points not covered by the region growing process are clustered and iteratively assigned to the closest tree.

The software 3DFin~\citep{laino-2024} implements a clustering-based stem segmentation approach for \ac{TLS} and \ac{PLS} data, where points corresponding to vertical structures are extracted from a point cloud and clustered with the DBSCAN algorithm. However, 3DFin does not delineate tree crowns accurately, so it does not provide full tree instance segmentation.

%% file: tab_0_related_work.tex
\begin{table*}[ht]
    \centering
    \begin{scriptsize}
    \begin{tabular}{
    L{0.26\linewidth-2\tabcolsep}
    L{0.15\linewidth-2\tabcolsep}
    L{0.22\linewidth-2\tabcolsep}
    M{0.14\linewidth-2\tabcolsep}
    M{0.07\linewidth-2\tabcolsep}
    M{0.07\linewidth-2\tabcolsep}
    M{0.09\linewidth-2\tabcolsep}}
    Reference & Name & Approach & Target Sensor & Open Source & Automated Tests & Doc\-u\-men\-ta\-tion \\ \midrule
    \cite{chang-2022} & - & 2D~DL + 3D~DL for sem. seg. & TLS & & & \cmark \\
    \cite{straker-2023} & - & 2D~DL & ULS & \cmark & & \cmark \\ \midrule
    \cite{xiang-2024} & ForAINet & 3D~DL (PointGroup) & ULS & \cmark & & \cmark \\
    \cite{wielgosz-2024} & Segment\-Any\-Tree & 3D~DL (PointGroup) & TLS, PLS, ULS, ALS & \cmark & & \cmark \\
    \cite{henrich-2024} & TreeLearn & 3D~DL (PointGroup) & TLS, PLS, ULS & \cmark & & \cmark \\
    \cite{xi-2025} & TreeisoNet & 3D~DL (SegFormer + PointGroup) & TLS, ULS, ALS & & & \cmark \\ 
    \cite{xiu-2025} & 3DPS-Net & 3D~DL (KPConv + prompt encoder) & TLS, ULS & \cmark & & \cmark \\
    \cite{xiang-2025} & Forest\-Former3D & 3D~DL (PointGroup + OneFormer) & TLS, PLS, ULS & \cmark & & \cmark\\ \midrule
    \cite{lowe-2021} & Ray\-Cloud\-Tools & graph-based & TLS, PLS & \cmark & \cmark & \\
    \cite{wang-2021} & - & graph-based & TLS & & & \cmark \\
    \cite{xi-2022} & treeiso & graph-based & TLS & \cmark & & \cmark \\
    \cite{xu-2023} & - & graph-based & TLS & & & \cmark \\
    \cite{wilkes-2023}, \cite{wielgosz-2023} & TLS2trees / Point2Tree & graph-based + DL for sem. seg. & TLS & \cmark & & \cmark \\ \midrule
    \cite{trochta-2017} & 3D~Forest & clustering & TLS & \cmark & & \cmark \\
    \cite{burt-2019} & treeseg & clustering \& region growing & TLS & \cmark & & \cmark \\
    \cite{bienert-2021} & - & clustering \& region growing & MLS & & & \cmark \\
    \cite{fu-2022} & - & clustering & TLS & & & \cmark \\
    \cite{tockner-2022}, \cite{gollob-2020} & treeX & clustering \& region growing & PLS & \cmark & & \cmark \\
    \cite{laino-2024} & 3DFin & clustering (no crown delineation) & TLS, PLS & \cmark & & \cmark
    \end{tabular}
    \end{scriptsize}
    \caption{Overview of existing tree instance segmentation methods. The methods are grouped into 2D~deep learning, 3D~deep learning, graph-based, and clustering- and region-growing-based methods. The methods within the same category are sorted by the date of their publication (DL = deep learning, sem. seg. = semantic segmentation). The column \enquote{Automated tests} indicates whether test suites implemented in code are available for the respective algorithm. This was determined by checking the respective code repositories, which was only possible for methods with open-source implementations. The column \enquote{Documentation} specifies whether a written description of the algorithm exists, either in the form of a scientific publication or another accessible online document.}
    \label{tab:related-work}
\end{table*}

%% file: sec_3_algorithm.tex
In this paper, we present a revised version of the treeX algorithm, which was originally proposed by \cite{tockner-2022} and \cite{gollob-2020}. The algorithm is an unsupervised, algorithmic method for segmenting tree instances in dense 3D~point clouds of forest environments. The input is a 3D~point cloud, defined by xyz-coordinates for each point. Optionally, reflectance intensity values may be included to improve the segmentation process. The output of the algorithm consists of tree instance IDs for each point (non-tree points are marked with a special ID), as well as lists of the detected tree's positions and diameters at breast height (DBHs)\acused{DBH}. As shown in Fig.~\ref{fig:algorithm-overview}, the algorithm consists of three main stages: (1)~construction of a digital terrain model, (2)~detection of tree stems, and (3)~delineation of tree crowns.

\input{fig_dtm_construction}
\input{tab_1_dtm_construction_parameters}

\subsection{Construction of a Digital Terrain Model}

In the first stage, a \ac{DTM} is reconstructed from the input point cloud (Fig.~\ref{fig:dtm-construction}). For this purpose, the point cloud is classified into terrain and non-terrain points using the unsupervised \ac{CSF} algorithm~\citep{zhang-2016}. The \ac{CSF} algorithm inverts the point cloud along the z-axis and simulates a cloth covering the inverted point cloud. Points within a certain distance from the simulated cloth surface are classified as terrain points.
Subsequently, the identified terrain points are down-sampled using voxel-based sampling, and a rasterized \ac{DTM} is derived. To this end, the \ac{DTM} values are interpolated using a k-nearest neighbors approach with inverse distance weighting similar to the \textit{grid\_terrain} method of the lidR package~\citep{roussel-2020}.
The terrain height $h(q)$ at grid position $q$ is computed using the following formula:
\begin{equation}
\label{eqn:dtm-rasterization}
    h(q) = \frac{1}{\sum \limits_{p \in \mathcal{N}(q, k)} w(q, p)} \cdot \sum_{p \in \mathcal{N}(q, k)} p_z \cdot w(q, p)
\end{equation}
where $\mathcal{N}(q, k)$ is the set of the $k$ terrain points closest to grid position $q$, $p_z$ is the z-coordinate of point $p$, and $w$ is an inverse distance weighting function with a hyperparameter $c$:
\begin{equation}
\label{eqn:dtm-rasterization-distance-weighting}
    w(q, p) = \frac{1}{||p_{xy} - q_{xy}||^c}
\end{equation}
The rasterized DTM is used in several steps of our algorithm to compute the vertical distance of points from the terrain surface, in the following referred to as height.
The height of a dedicated point is calculated by performing a bilinear interpolation between the four nearest grid nodes of the rasterized \ac{DTM} surrounding the point.

All parameters of the \ac{DTM} construction stage and their default values are shown in Table~\ref{tab:dtm-construction-parameters}. The default values were selected experimentally by testing the algorithm on various sample point clouds and visually inspecting the results.

\input{fig_algorithm_overview}

\subsection{Detection of Tree Stems}

Tree stems are detected by extracting a stem layer (1) and identifying high-density clusters among the corresponding points (2). Afterwards, false positives are filtered out (3), and finally, stem positions are determined and corresponding \acp{DBH} are derived (4).

\paragraph{(1) Stem layer extraction} In this step, a horizontal layer is extracted from the 3D~point cloud, containing all points whose distance to the \ac{DTM} is between a minimum height \(h_{min}\) and a maximum height \(h_{max}\) (Fig.~\ref{fig:stem-layer}). To reduce the computational cost of the following two processing steps (DBSCAN clustering and filtering of clusters), the points in this layer are down-sampled using voxel-based sampling with a default voxel size of \num{0.015}\,m (Table~\ref{tab:stem-detection-parameters}).

\paragraph{(2) DBSCAN clustering} To identify potential tree stems, points within the stem layer are clustered using the DBSCAN algorithm~\citep{ester-1996}. The DBSCAN algorithm is a density-based method that identifies clusters of closely packed points and classifies each point either as core point, border point, or noise~(Fig.~\ref{fig:dbscan}). A point is considered a core point if it has at least~$MinPts$ neighbors within a radius~$\epsilon$ ($\epsilon$-neighborhood), both of which are user-defined parameters. Border points are those that lie within the $\epsilon$-neighborhood of at least one core point, while not being core points themselves. Points that are neither core points nor boarder points are labeled as noise and are not assigned to any cluster.

To detect tree stems, the DBSCAN algorithm is applied in two stages: First, it is applied in 2D using only the xy-coordinates of the points (Fig.~\ref{fig:dbscan-2d}). Since tree stems tend to form dense clusters when projected onto the xy-plane, the 2D~DBSCAN parameters are set to $\epsilon =$ \num{0.025}\,m and $MinPts = 90$ for \ac{TLS} and \ac{PLS} data, and to $\epsilon =$ \num{0.07}\,m and $MinPts = 15$ for \ac{ULS} data. However, in areas with high vegetation density, closely spaced stems or branches may be merged into a single 2D~cluster.
To further split such clusters, a second DBSCAN clustering step is applied to each of the 2D~clusters, performing a 3D~clustering using xyz-coordinates (Fig.~\ref{fig:dbscan-3d}). In this stage,  \(\epsilon\) is increased to \num{0.1}\,m for \ac{TLS} and \ac{PLS} data and to \num{0.3}\,m for \ac{ULS} data, and \(MinPts\) is reduced to \num{15} and \num{1}, respectively, to avoid over-segmentation.

\paragraph{(3) Filtering of clusters} With the goal of discarding clusters that do not represent tree stems, the following filtering rules are applied to the clusters obtained with the DBSCAN algorithm: 

\begin{enumerate}
    \item Clusters containing fewer than $N_{min}$ points are discarded.
    \item Clusters whose vertical extent is less than $\Delta H_{min}$ are discarded.
    \item  If the input point cloud includes reflectance intensity values, the \num{80}\,\% percentile of the intensity values of the points in a cluster is calculated. Since stem points tend to have greater reflectivity than leaf points, clusters with a percentile smaller than $I_{min}$ are discarded.
    \item The remaining clusters are further filtered based on the assumption that the diameter of tree stems changes gradually with height. To filter out clusters with significant diameter variations at different heights, horizontal layers are extracted, and circles are fitted to the points within each layer. Clusters for which the radii of the fitted circles vary more than a certain threshold are discarded. This filtering procedure is described in detail in the following steps:

    \begin{enumerate}
        \item Initially, $N_{l}$ horizontal layers are extracted from each cluster, and the points within each layer are collected. The first layer starts at a height of $h_0$ above the terrain, and the layers have a height of $h_l$ and a vertical overlap of $o_l$. To determine whether a point is contained in a layer, its normalized z-coordinate is used, which is obtained by computing the \ac{DTM} height at the cluster centroid position and subtracting it from the z-coordinate of the point.
        \item Next, a circle is fitted to the xy-coordinates of the points within each layer (Fig.~\ref{fig:circle-fitting}). A circle fitting method based on \ac{RANSAC}~\citep{fischler-1981} is used for this purpose. Details on the circle fitting method and its parametrization are provided in \ref{sec:circle-fitting-details}.
        To ensure that only circles corresponding to realistic stem diameters are detected, the circle fitting method is constrained to detect circles with diameters within a user-specified range between $\diameter_{min}$ and $\diameter_{max}$.
        Optionally, the circles can also be filtered based on their \ac{CCI}~\citep{krisanski-2020}.
        \item Finally, the fitted circles are used to filter the clusters.
        Clusters for which circle fitting was successful in less than \(N_{\mathrm{sample}}\) layers, e.g., due to insufficient point density in most layers, are discarded.
        Each remaining cluster is evaluated by identifying the subset of \(N_{\mathrm{sample}}\) layers for which the diameters of the fitted circles exhibit the smallest standard deviation.
        To this end, all possible combinations of \(N_{\mathrm{sample}}\) layers are tested.  
        If the standard deviation of the circle diameters in the best subset does not exceed a predefined threshold \(\sigma_{\max}^r\), the cluster is considered valid and retained.
        Optionally, the remaining clusters can be further filtered by evaluating the spatial consistency of the fitted circles:
        For the layer subset with the smallest standard deviation of circle diameters, the standard deviation of the circle center positions is computed.
        If this value exceeds a threshold \(\sigma_{\max}^c\), the cluster is discarded.  
        This filtering based on circle center positions is only recommended in scenarios where tree stems are expected to grow approximately vertically and is therefore disabled by default.
    \end{enumerate}

\end{enumerate}

\input{fig_dbscan}
\input{fig_stem_clustering}

All clusters that pass the above filter steps are regarded as valid stem detections.

\input{tab_2_stem_detection_parameters}

\paragraph{(4) Calculation of stem positions and diameters}

The objective of this step is to accurately estimate the stem center position and the diameter at breast height for each detected stem cluster.
For each cluster, only the subset of layers with the smallest standard deviation of circle diameters identified in the previous filtering step is considered.
For each layer in this subset, stem diameters are re-estimated using a \ac{GAM}: First, all points within a buffer region of width \(b\) around the outline of the fitted circle are extracted.
These points are then normalized by subtracting the center coordinates of the fitted circle and transformed into a polar coordinate system, where each point is represented by an angle and a radius.
A \ac{GAM} is then fitted to model the radius as a function of the angle. The fitted model is used to predict the stem radius at 1-degree intervals (Fig.~\ref{fig:gam-fitting}). These predictions are converted back into Cartesian coordinates and treated as vertices of a polygon approximating the stem cross-section. The square root of the area of this polygon is used as the stem diameter estimate for that layer.
In cases of low point density or severe occlusions, the fitted \ac{GAM} may deviate strongly from the underlying stem shape. To exclude such cases, the difference between the maximum and minimum predicted radius values is calculated. If this difference exceeds a threshold \(\Delta r_{\max}\), the \ac{GAM} fit is considered invalid, and the diameter estimated from the previously fitted circle is used instead.

To determine the stem diameter at breast height, a linear regression model is fitted to the \ac{GAM}-based diameter estimates across the selected layers. The layer heights serve as the independent variable, and the corresponding \ac{GAM}-based diameter estimates as the dependent variable. The predicted diameter at a height of \num{1.3}\,m is taken as the final \ac{DBH} estimate.
The stem center position at breast height is estimated similarly, by fitting separate linear models to the x- and y-coordinates of the circle centers across layers and predicting their values at \num{1.3}\,m.
\\
\\
All parameters of the stem detection stage of our algorithm, along with their default values, are summarized in Table~\ref{tab:stem-detection-parameters}.

\input{fig_shape_fitting}

\subsection{Tree Crown Delineation}

While the previous stage identified the positions and diameters of individual tree stems, this stage aims to determine the complete sets
of points that represent each corresponding tree. In particular, this involves segmenting the canopy points into individual tree crowns. To accomplish this, the region growing method proposed by~\citet{tockner-2022} is employed. The method starts by selecting a set of seed points for each tree based on the previously detected stem positions and diameters~(1). As shown in Fig.~\ref{fig:region-growing}, it then iteratively expands each tree segment by incorporating neighboring points of the seed points, which are subsequently used as new seed points in the next iteration (2). To improve computational efficiency and ensure consistent point density, the point cloud is down-sampled using voxel-based sampling prior to the region growing process.

\paragraph{(1) Initial seed point selection} The initial seed points for a given tree are extracted from a cylindrical subsection of the point cloud centered at the stem position.
The diameter of this subsection is set to $DBH \cdot f_{seed}$, where $f_{seed}$ is a user-defined scaling factor. To ensure that a sufficient number of seed points is selected for small trees, the diameter of the cylinder is constrained to a minimum value of $\diameter_{min}^{seed}$. Vertically, the cylinder is centered at breast height (\num{1.3}\,m) and extends by $0.5 \cdot h_{\mathrm{seed}}$ above and below this level. The terrain height at the stem position is used as reference for vertical alignment.

\input{fig_region_growing}
\input{tab_3_crown_delineation_parameters}

\paragraph{(2) Region growing loop}
The region growing is performed iteratively. In each iteration, for every tree, a list of current seed points is processed: neighboring points around each seed point are identified using a radius search and are assigned to the corresponding tree if they have not already been assigned to a tree. If an unassigned point lies within the neighborhoods of seed points from multiple trees, it is assigned to the tree whose seed point is closest. To promote upward expansion of tree crowns, the z-coordinates of all points are divided by a scaling factor $f_z$ prior to region growing.

Choosing an appropriate search radius is critical to the performance and accuracy of the region growing process. A search radius that is too small can result in slow expansion, while a search radius that is too large increases the risk of segmentation errors, as tree segments may erroneously expand to points from neighboring trees. To balance these aspects, the initial search radius is set equal to the voxel size used in the down-sampling step. The radius is then dynamically adjusted after each iteration based on two metrics: the total assignment ratio, defined as the proportion of newly assigned points relative to the total number of unassigned points at the start of the iteration, and the tree assignment ratio, defined as the proportion of trees that received new points during the iteration. If either of these metrics falls below user-defined thresholds, $t_{total}$ and $t_{trees}$, respectively, the search radius is doubled at the start of the next iteration. Conversely, if the search radius has remained unchanged for $\Delta t_{\downarrow}$ consecutive iterations, it is halved to mitigate the risk of segmentation errors. This reduction is applied only if the current search radius exceeds the voxel size used in the initial down-sampling.

The selection of seed points at the beginning of each iteration is done as follows: 
In the first iteration, seed points are selected using the stem positions and diameters, as described in step (1). In subsequent iterations, the points newly assigned in the previous iteration are used as seed points if the search radius has remained constant or decreased compared to the previous iteration. If the search radius has increased, all currently assigned points are used as seed points.

Since stem bases may be classified as terrain points in the \ac{DTM} construction stage of our algorithm, both terrain and non-terrain points are considered in the region growing process. However, to prevent tree instances from incorporating large terrain patches, terrain points are only assigned to a tree instance if their cumulative search distance to an initial seed point from the first region growing iteration is below a threshold \(d_{terrain}\).

The region growing loop terminates when either the list of seed points is empty, a maximum search radius has been reached, or a maximum number of iterations has been reached.
After the region growing loop is completed, the resulting tree instance labels are upsampled to the full resolution of the point cloud by propagating the labels of the down-sampled points to all points within their corresponding voxels.
\\
\\
All parameters of the tree crown delineation stage of our algorithm and their default values are summarized in Table~\ref{tab:crown-segmentation-parameters}.

\subsection{Parameter Presets}

While the original treeX algorithm was developed with a focus on \ac{PLS} data, we aim to support a broader range of point cloud sources in our revised version. The choice of suitable settings for the stem detection stage of the algorithm depends on the density of the input point cloud. To account for varying densities, we provide two parameter presets: The first preset, referred to as the \ac{TLS} preset, is intended for point clouds with high densities in the stem layer, e.g., those acquired with \ac{TLS} or \ac{PLS}. The second, the \ac{ULS} preset, targets lower-density \ac{ULS} data. Compared to the \ac{TLS} preset, the \ac{ULS} preset adjusts several parameters of the stem detection stage: it increases the vertical extent of the horizontal layer considered for stem detection, lowers the density thresholds for DBSCAN clustering, reduces the minimum required number of points per cluster, and adjusts the parameters of the circle fitting (Table~\ref{tab:stem-detection-parameters}). The parameters of the \ac{DTM} construction stage and the tree crown delineation stages are the same for both parameter presets.

\subsection{Implementation Details}

The algorithm was implemented in Python and integrated into the open-source package pointtree~\citep{burmeister-2025c}.\footnote{Repository of the pointtree package: \url{https://github.com/ai4trees/pointtree}}
NumPy arrays~\citep{harris-2025,harris-2020}\footnote{Repository of the NumPy package: \url{https://github.com/numpy/numpy}} are used as the primary data structure for both input and output, as these are widely used. Besides NumPy, the main dependencies of our implementation include the Python packages scikit-learn~\citep{pedregosa-2025,pedregosa-2011},\footnote{Repository of the scikit-learn package: \url{https://github.com/scikit-learn/scikit-learn}} scipy~\citep{virtanen-2025,virtanen-2020},\footnote{Repository of the scipy package: \url{https://github.com/scipy/scipy}} pointtorch~\citep{burmeister-2025b},\footnote{Repository of the pointtorch package: \url{https://github.com/ai4trees/pointtorch}}  CSF~\citep{zhang-2025},\footnote{Repository of the CSF package: \url{https://github.com/jianboqi/CSF}} circle-detection~\citep{burmeister-2025a},\footnote{Repository of the circle-detection package: \url{https://github.com/josafatburmeister/circle_detection}} and pyGAM~\citep{serven-2025}.\footnote{Repository of the pyGAM package: \url{https://github.com/dswah/pyGAM}}
To optimize the performance of the algorithm, computationally intensive steps were implemented in C++ and integrated into the Python code via pybind11~\citep{jakob-2025}.\footnote{Repository of pybind11: \url{https://github.com/pybind/pybind11}} The main dependencies of the C++ implementation are Eigen~\citep{jacob-2025}\footnote{Repository of the Eigen library: \url{https://gitlab.com/libeigen/eigen}} and nanoflann~\citep{blanco-claraco-2025}.\footnote{Repository of the nanoflann library: \url{https://github.com/jlblancoc/nanoflann}}
The source code of our implementation was documented extensively using Sphinx~\citep{brandl-2025}.\footnote{Repository of the sphinx package: \url{https://github.com/sphinx-doc/sphinx}}
To ensure the correctness of the implementation, extensive unit tests were developed, achieving a test coverage of over \num{90}\,\%.

%% file: fig_dtm_construction.tex
\begin{figure}[ht]
    \centering
    \begin{subfigure}[t]{0.49\linewidth}
    \centering
    \includegraphics[width=\linewidth]{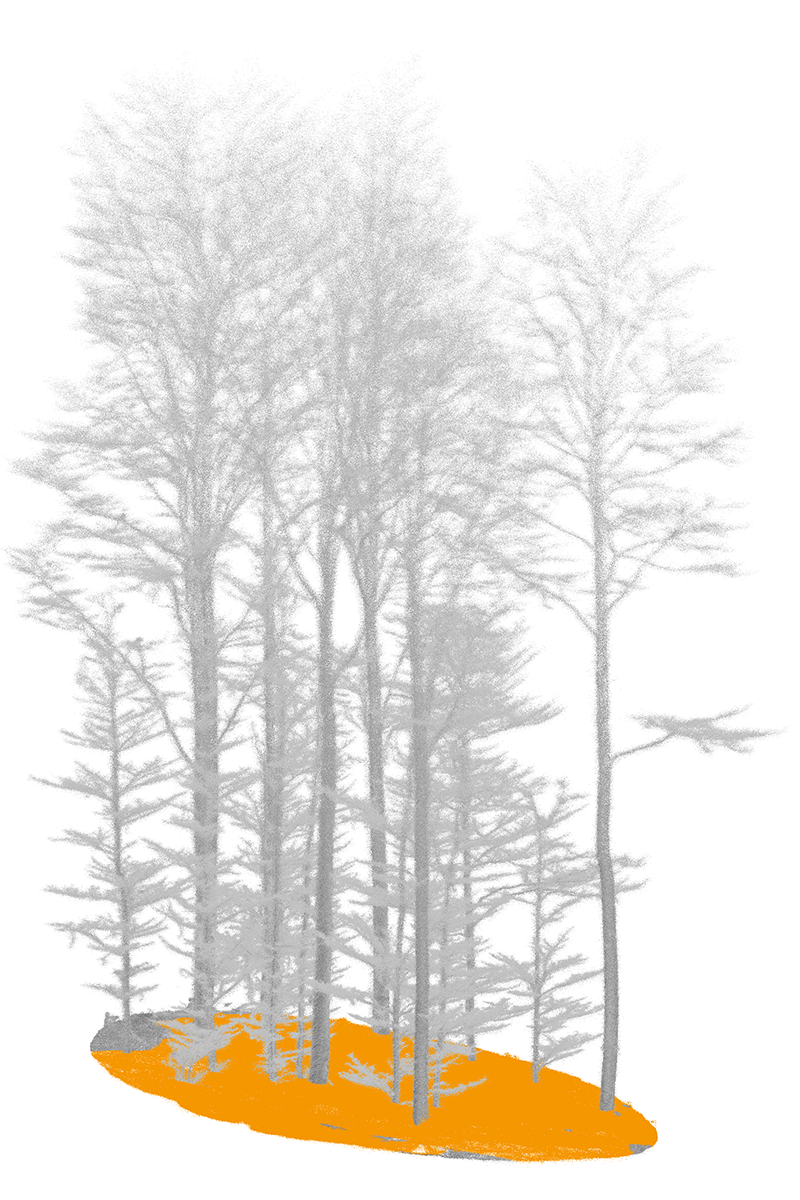}
    \caption{Terrain classification results based on the CSF algorithm (points classified as terrain are shown in orange, non-terrain points in gray)}
    \label{fig:dtm-construction-csf}
    \end{subfigure}
    \begin{subfigure}[t]{0.49\linewidth}
    \centering
    \includegraphics[width=\linewidth]{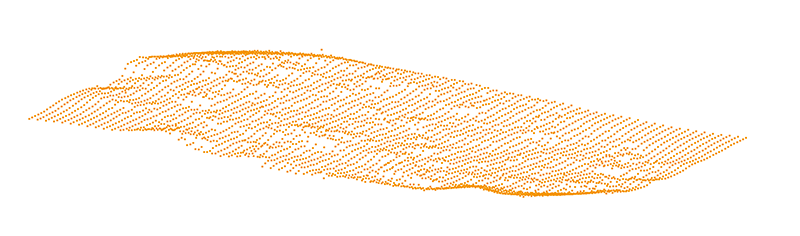}
    \caption{Rasterized digital terrain model (DTM)}
    \label{fig:dtm-construction-dtm}
    \end{subfigure}
    \caption{Example results of the digital terrain model construction method used in our algorithm.}
    \label{fig:dtm-construction}
\end{figure}

%% file: tab_1_dtm_construction_parameters.tex
\begin{table}[ht]
    \centering
    \begin{scriptsize}
    \renewcommand{\arraystretch}{1.3}
    \begin{tabular}{L{0.23\linewidth-2\tabcolsep}L{0.65\linewidth-2\tabcolsep}M{0.12\linewidth-2\tabcolsep}}
         Parameter & Description & Default Value  \\ \toprule
         CSF cloth resolution & Distance between neighboring nodes of the simulated cloth in the CSF algorithm & \num{0.5}\,m \\
         CSF cloth rigidness & Rigidness of the simulated cloth in the CSF algorithm (min.~=~1, max.~=~3) & 2 \\
         CSF max. iterations & Maximum number of iterations in the CSF algorithm & 500 \\
         CSF steep slope fit factor & Whether to include a post-processing step in the CSF algorithm that handles steep slopes  & False \\
         Terrain classification threshold & Maximum distance from the simulated cloth surface up to which points are classified as terrain points & \num{0.5}\,m \\
         Tree classification threshold & Minimum distance from the terrain surface from which points are classified as tree points (must be $\geq$ terrain classification threshold) & \num{0.5}\,m \\
         DTM voxel size & Voxel size for voxel-based down-sampling of the terrain points before constructing the rasterized DTM & \num{0.05}\,m \\
         DTM resolution & Distance between neighboring nodes in the rasterized DTM & \num{0.25}\,m \\
         k & Number of neighboring terrain points considered in the construction of the rasterized DTM, as defined in Eq.~\ref{eqn:dtm-rasterization} & 400 \\
         c & Power of the inverse distance weighting function used in the construction of the rasterized DTM, as defined in Eq.~\ref{eqn:dtm-rasterization-distance-weighting} & 1
    \end{tabular}
    \end{scriptsize}
    \caption{Parameters of the digital terrain model construction stage of our algorithm.}
    \label{tab:dtm-construction-parameters}
\end{table}

%% file: fig_algorithm_overview.tex
\newenvironment{tightitemize}{
    \begin{itemize}[topsep=3pt, itemsep=-2pt, leftmargin=8pt, labelsep=2pt]
    \renewcommand{\labelitemi}{\tiny\textbullet}
}{
  \end{itemize}
}

\begin{figure*}
    \centering
    \includegraphics{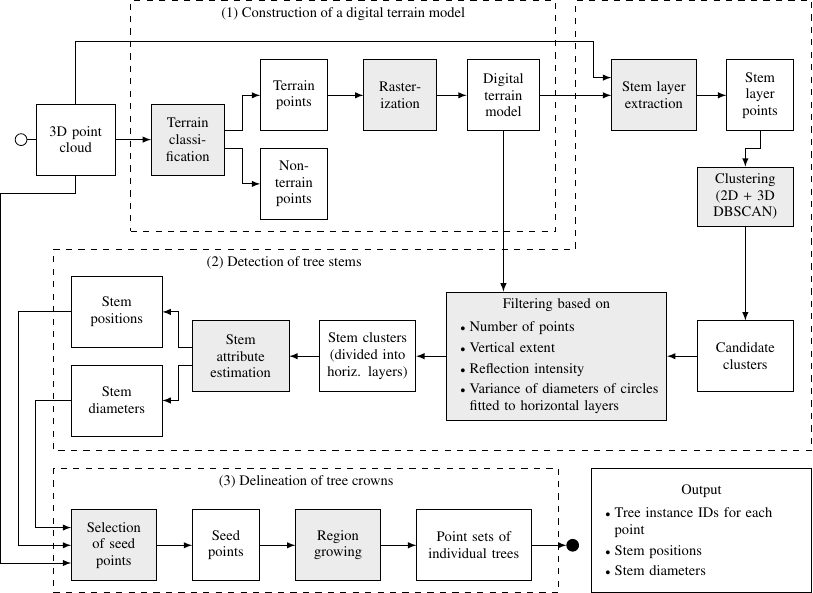}
    \caption{Overview of the treeX algorithm (updated version as proposed in this work). White boxes represent data, and gray boxes represent data processing steps.}
    \label{fig:algorithm-overview}
\end{figure*}

%% file: fig_dbscan.tex
\begin{figure}
\centering
\includegraphics{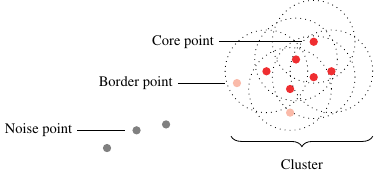}
\caption[Core points, border points, and noise points as defined in the DBSCAN algorithm]{Core points, border points, and noise points as defined in the DBSCAN algorithm~\citep{ester-1996}. The dotted circles represent the $\epsilon$-neighborhoods of the core points, and the clustering shown was generated with $MinPts = 3$.}
\label{fig:dbscan}
\end{figure}

%% file: fig_stem_clustering.tex
\begin{figure}
    \centering
    \begin{subfigure}[t]{0.23\linewidth}
    \centering
    \includegraphics[trim={0 0 0 25px}, clip, width=\linewidth]{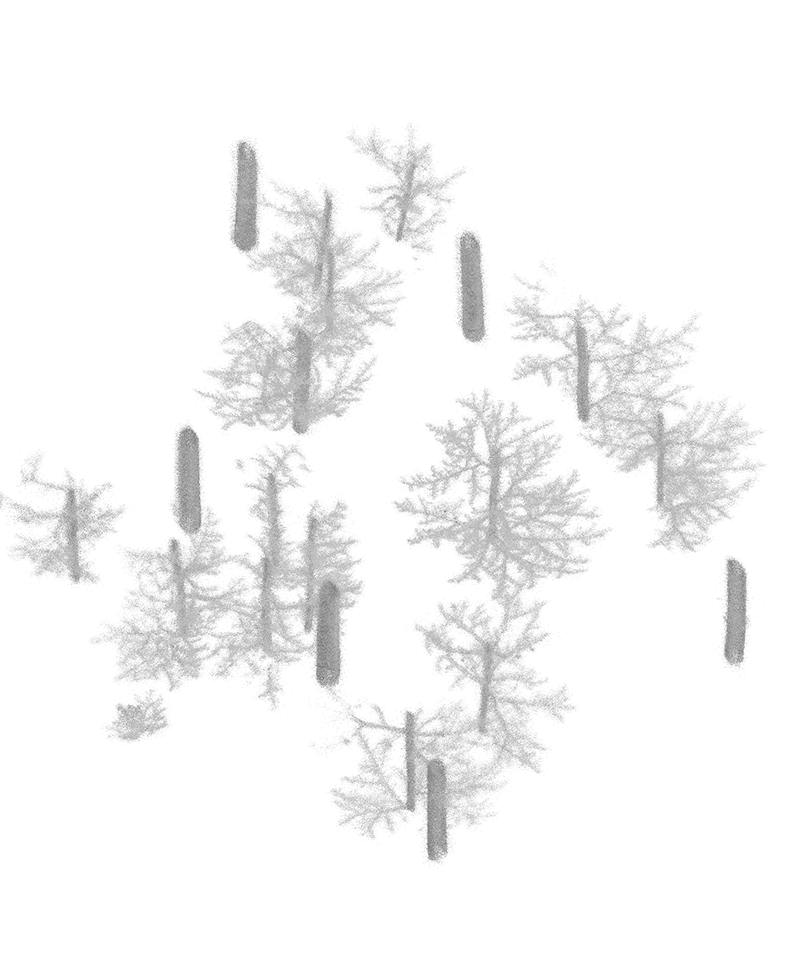}
    \caption{Stem layer points}
    \label{fig:stem-layer}
    \end{subfigure}
    \begin{subfigure}[t]{0.23\linewidth}
    \centering
    \includegraphics[trim={0 0 0 25px}, clip, width=\linewidth]{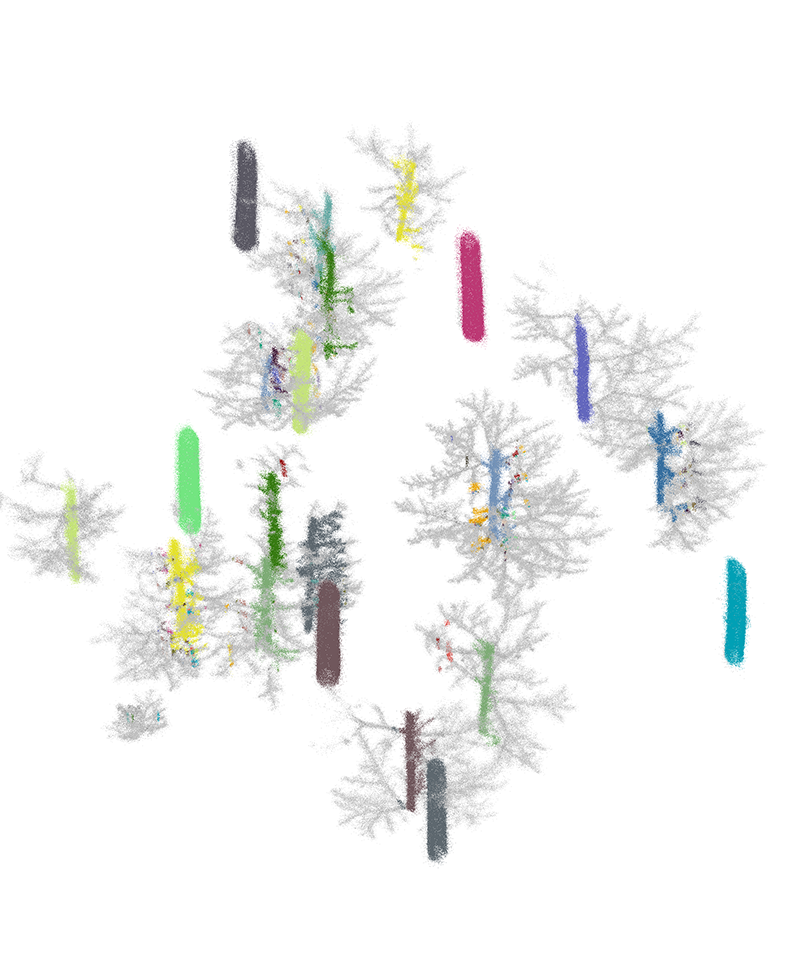}
    \caption{Clusters identified using 2D DBSCAN}
    \label{fig:dbscan-2d}
    \end{subfigure}
    \begin{subfigure}[t]{0.23\linewidth}
    \centering
    \includegraphics[trim={0 0 0 25px}, clip, width=\linewidth]{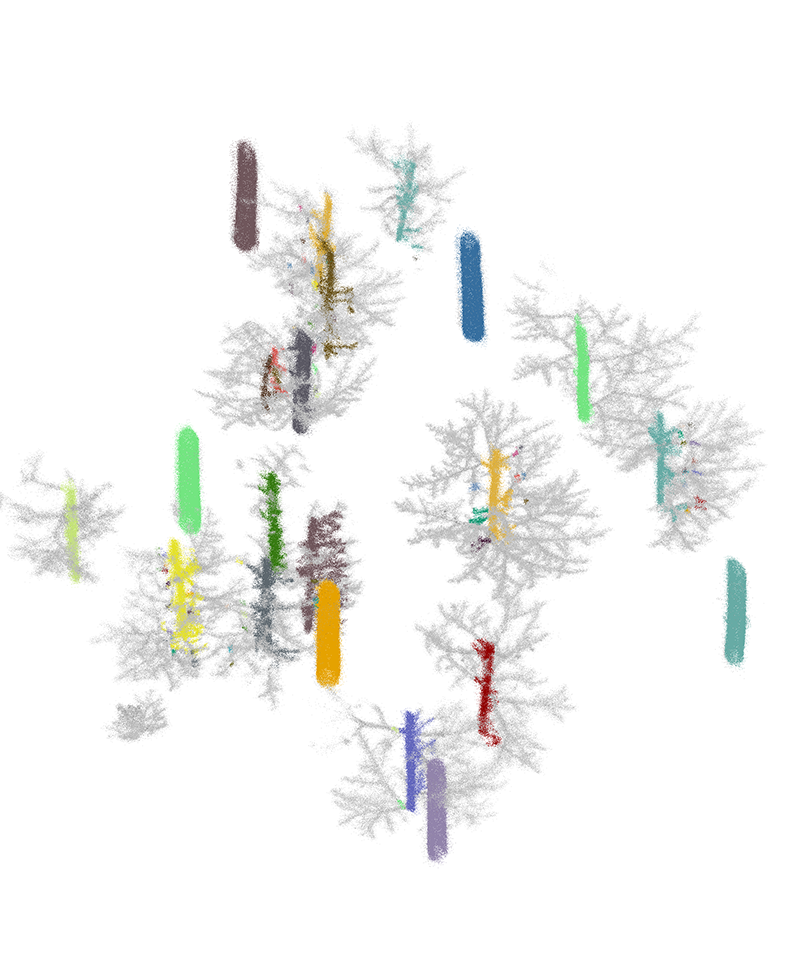}
    \caption{Refined clusters after applying 3D DB\-SCAN}
    \label{fig:dbscan-3d}
    \end{subfigure}
    \begin{subfigure}[t]{0.23\linewidth}
    \centering
    \includegraphics[trim={0 0 0 25px}, clip, width=\linewidth]{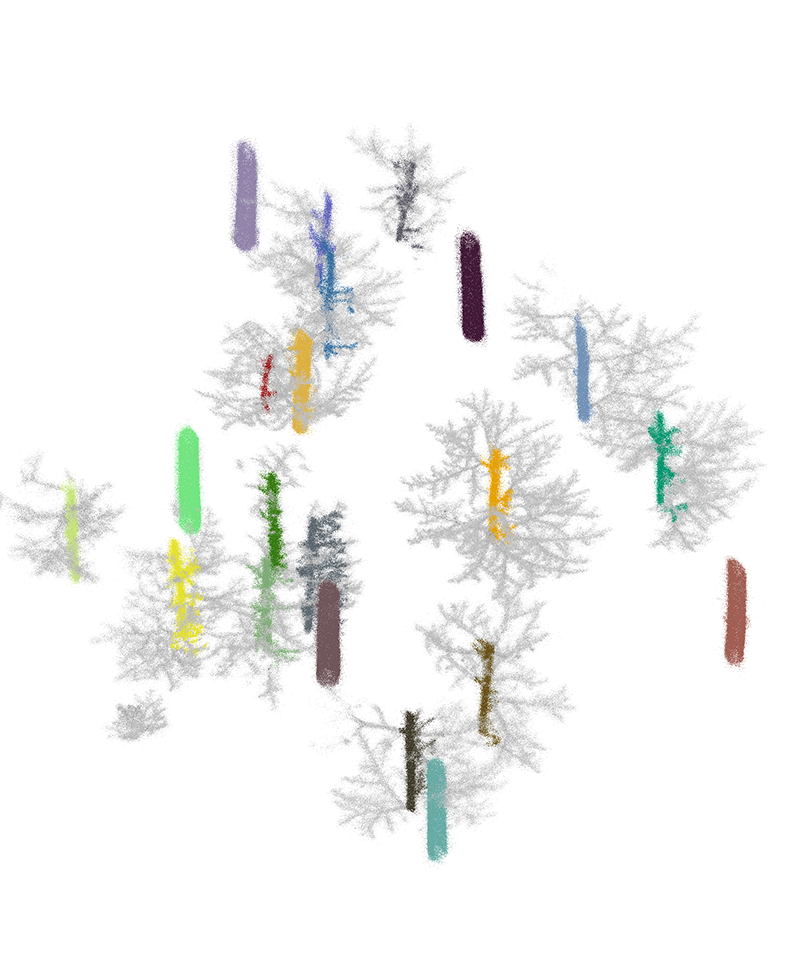}
    \caption{Final clusters}
    \label{fig:final-stems}
    \end{subfigure}
    \caption{Example results of the stem detection stage of our algorithm.}
    \label{fig:stem-clustering}
\end{figure}

%% file: tab_2_stem_detection_parameters.tex
\begin{table}[ht!]
    \centering
    \begin{scriptsize}
    \renewcommand{\arraystretch}{1.3}
    \begin{tabular}{L{0.23\linewidth-2\tabcolsep}L{0.53\linewidth-2\tabcolsep}M{0.12\linewidth-2\tabcolsep}M{0.12\linewidth-2\tabcolsep}}
         Parameter & Description & Default Value TLS Preset & Default Value ULS Preset  \\ \toprule
         $h_{min}$ & Lower boundary of the stem layer with respect to the height above the DTM & \num{1}\,m & \num{1}\,m \\
         $h_{max}$ & Upper boundary of the stem layer with respect to the height above the DTM & \num{4}\,m & \num{5}\,m \\
         Stem detection voxel size & Voxel size for voxel-based down-sampling of the points within the stem layer & \num{0.015}\,m & \num{0.015}\,m \\
         $\epsilon_{2D}$ & $\epsilon$ parameter of the DBSCAN algorithm used for 2D clustering & \num{0.025}\,m & \num{0.07}\,m \\
         $MinPts_{2D}$ & $MinPts$ parameter of the DBSCAN algorithm used for 2D clustering &  \num{90} & \num{15} \\
         $\epsilon_{3D}$ & $\epsilon$ parameter of the DBSCAN algorithm used for 3D clustering & \num{0.1}\,m & \num{0.3}\,m \\
         $MinPts_{3D}$ & $MinPts$ parameter of the DBSCAN algorithm used for 3D clustering & \num{15} & \num{1} \\
         $N_{min}$ & Minimum number of points that a cluster must contain in order not to be filtered out & \num{300} & \num{20}
         \\
         $\Delta H_{min}$ & Minimum vertical extent that a cluster must have in order not to be filtered out & \num{1.5}\,m & \num{1.5}\,m
         \\
         $I_{min}$ & Minimum threshold that the \num{80}\,\% percentile of the reflection intensity values of a cluster must reach in order not to be filtered out (only applicable if the input point cloud includes reflectance intensity values) & \num{6000}  & \num{6000}
         \\
         $\diameter_{min}$ & Minimum stem diameter (used for the parameterization of the circle fitting) & \num{0.02}\,m & \num{0.02}\,m
         \\
         $\diameter_{max}$ & Maximum stem diameter (used for the parameterization of the circle fitting) & \num{1}\,m & \num{1}\,m
         \\
         \(N_{l}\) & Number of horizontal layers per cluster used for circle fitting & 15 & 4
         \\
         \(h_0\) & Height above the DTM of the first horizontal layer used for circle fitting & \num{1}\,m & \num{1}\,m
         \\
         \(h_{l}\) & Height of the horizontal layers used for circle fitting & \num{0.225}\,m & \num{1.4}\,m
         \\
         \(o_{l}\) & Vertical overlap between adjacent horizontal layers used for circle fitting & \num{0.025}\,m & \num{0.4}\,m
         \\
         \(\mathcal{S}_{\min}\) & Minimum goodness-of-fit value for retaining a circle during fitting & \num{100} & \num{5} \\
         \(s\) & Bandwidth parameter for computing the goodness-of-fit in circle fitting; this value is also used as the error tolerance in RANSAC-based circle fitting & \num{0.01}\,m & \num{0.03}\,m
         \\
         \(N_{\min}^{circ}\) & Minimum number of points required for circle fitting & 15 & 3 \\
         \(CCI_{\min}\) & Minimum circular completeness index for retaining a circle during fitting (can be set to \enquote{None} to disable this filtering step) & 0.3 & 0.3
         \\
         \(N_{sample}\) & Number of horizontal layers to consider in each sample for calculating the standard deviation of the diameters of the fitted circles & \num{6} & \num{2}
         \\
         \(\sigma_{max}^{\diameter}\) & Threshold for filtering the clusters based on the standard deviation of the diameters of the fitted circles & \num{0.04}\,m & \num{0.1}\,m
         \\
         $\sigma_{max}^c$ & Threshold for filtering the clusters based on the standard deviation of the center positions of the fitted circles (can be set to \enquote{None} to disable this filtering step) & None & None \\
         \(b\) & Width of the buffer created around the outline of the fitted circles in order to determine the input points for the GAM-based stem diameter estimation & \num{0.03}\,m & \num{0.03}\,m \\
         \(\Delta r_{\max}\) & Maximum allowed difference between per-angle stem radii predicted by a GAM, beyond which the fitted model is considered invalid & \num{0.3}\,m & \num{0.3}\,m
    \end{tabular}
    \end{scriptsize}
    \caption{Parameters of the stem detection stage of our algorithm.}
    \label{tab:stem-detection-parameters}
\end{table}

%% file: fig_shape_fitting.tex
\begin{figure}
    \centering
    \begin{subfigure}[t]{0.49\linewidth}
    \centering
    \includegraphics[width=0.5\linewidth]{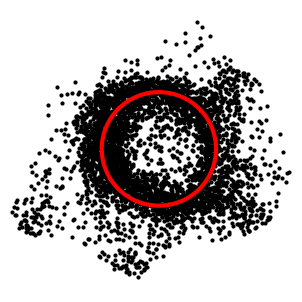}
    \caption{Circle fitting}
    \label{fig:circle-fitting}
    \end{subfigure}
    \begin{subfigure}[t]{0.49\linewidth}
    \centering
    \includegraphics[width=0.5\linewidth]{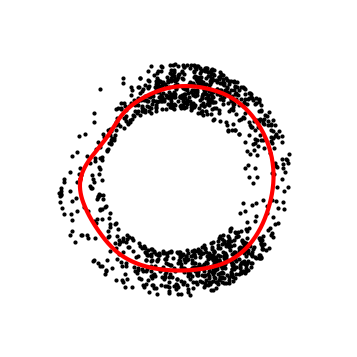}
    \caption{GAM fitting}
    \label{fig:gam-fitting}
    \end{subfigure}
    \caption{Example results of the shape fitting methods used in the stem detection stage of our algorithm.}
    \label{fig:shape-fitting}
\end{figure}

%% file: fig_region_growing.tex
\begin{figure}[ht]
    \centering
    \begin{subfigure}[t]{0.49\linewidth}
    \centering
    \includegraphics[width=0.9\linewidth]{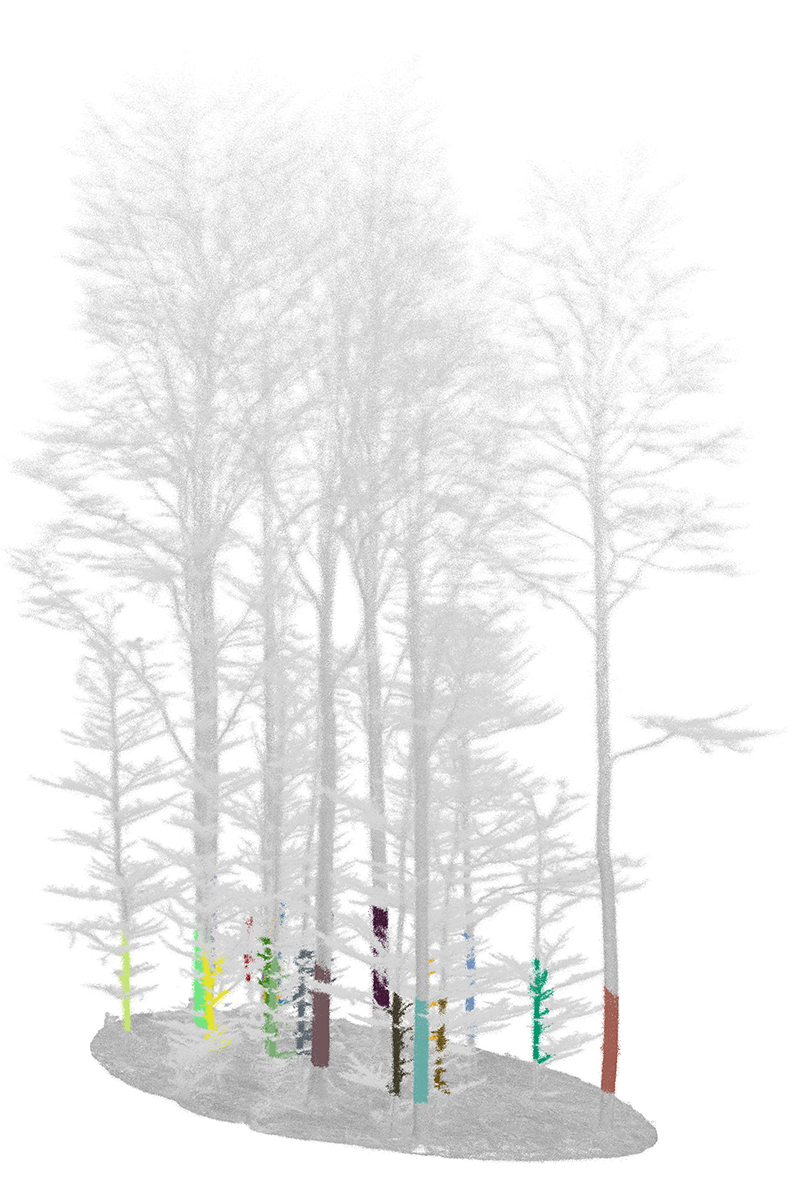}
    \caption{Initial seed points}
    \label{fig:region-growing-0}
    \end{subfigure}
    \begin{subfigure}[t]{0.49\linewidth}
    \centering
    \includegraphics[width=0.9\linewidth]{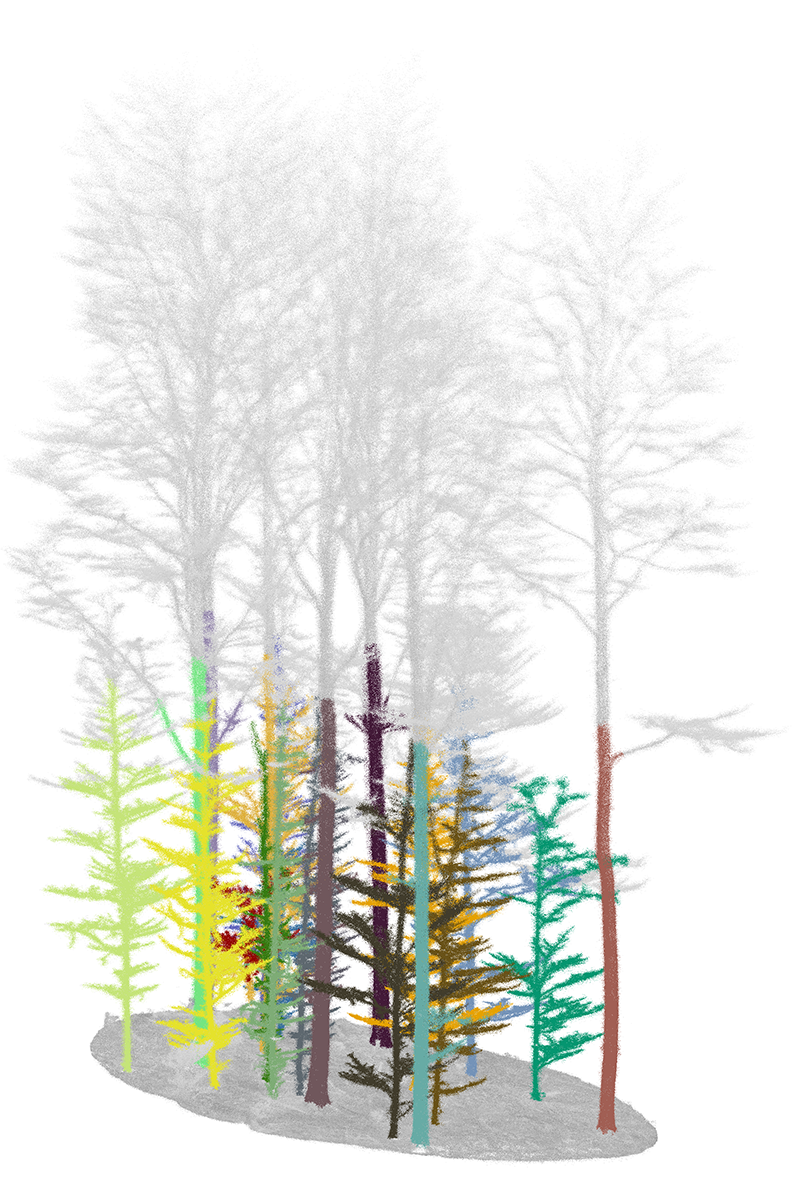}
    \caption{100 iterations}
    \label{fig:region-growing-100}
    \end{subfigure}
    \begin{subfigure}[t]{0.49\linewidth}
    \centering
    \includegraphics[width=0.9\linewidth]{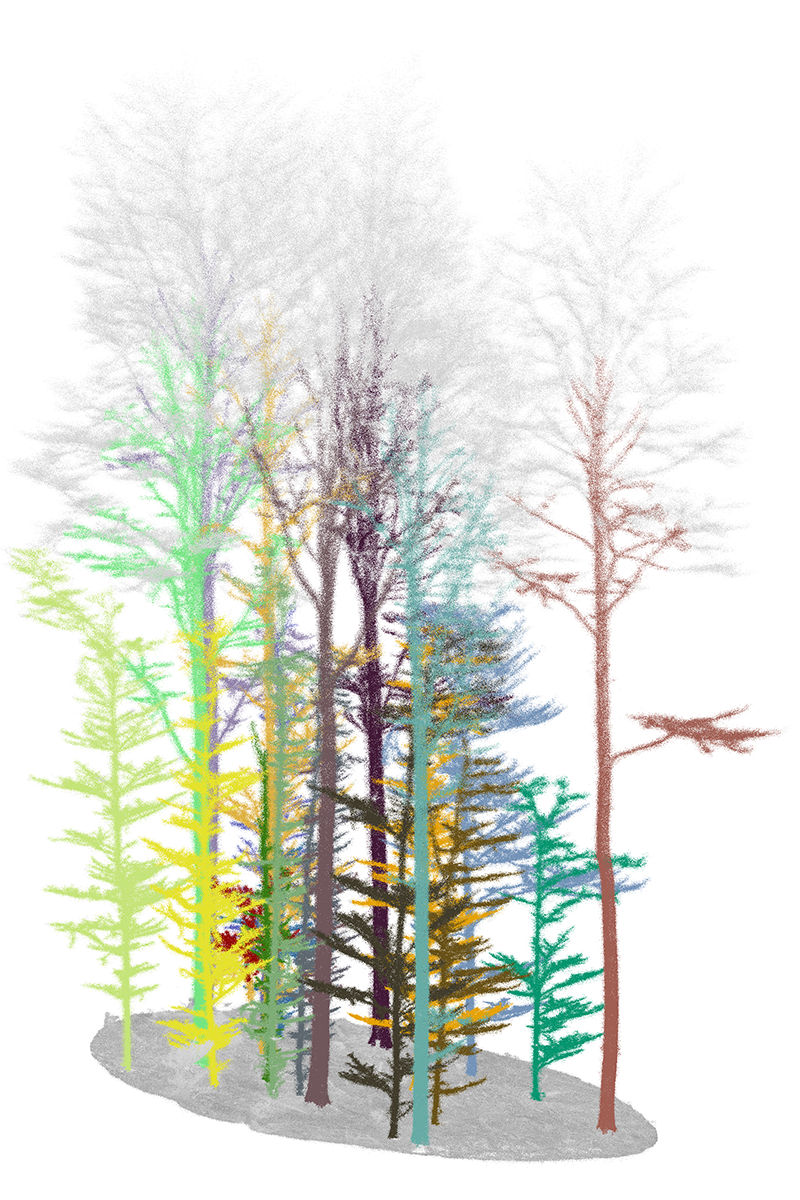}
    \caption{250 iterations}
    \label{fig:region-growing-250}
    \end{subfigure}
    \begin{subfigure}[t]{0.49\linewidth}
    \centering
    \includegraphics[width=0.9\linewidth]{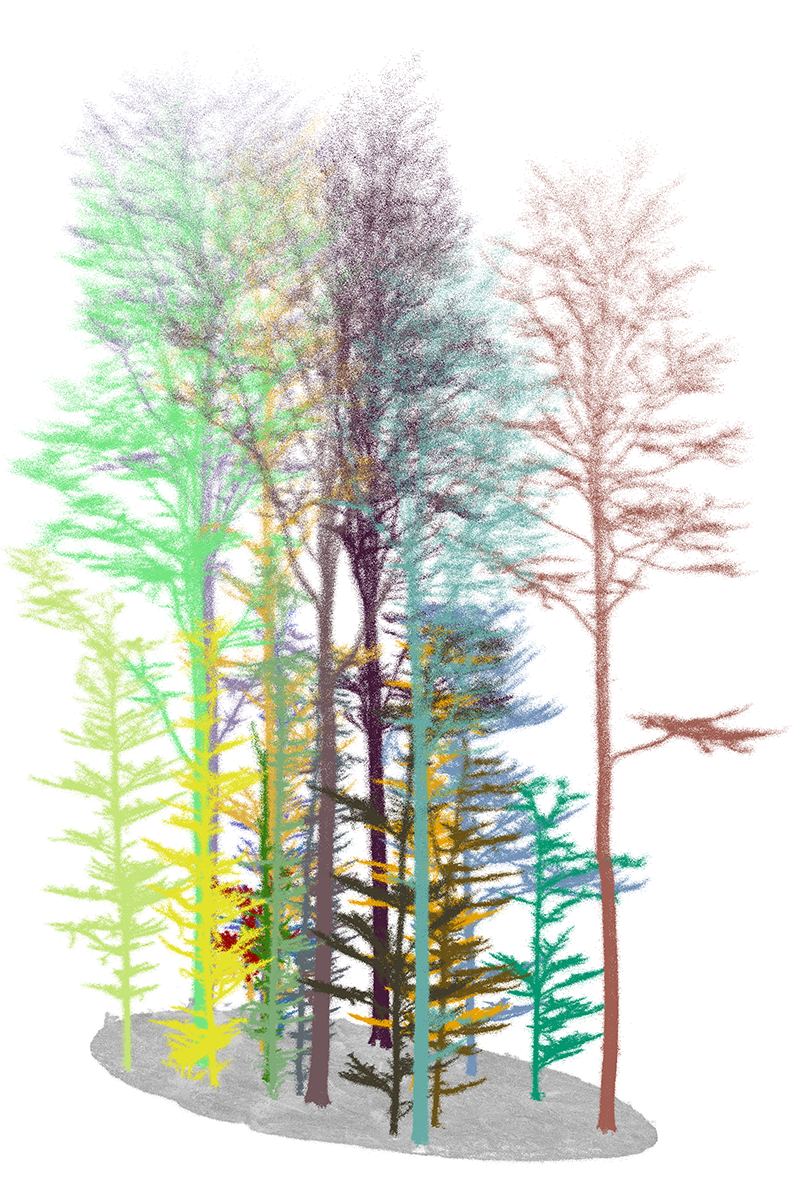}
    \caption{500 iterations}
    \label{fig:region-growing-1000}
    \end{subfigure}
    \caption{Intermediate results of the region growing method used for tree crown delineation in our algorithm after different numbers of iterations (different colors represent different trees, points not assigned to trees are shown in gray).}
    \label{fig:region-growing}
\end{figure}

%% file: tab_3_crown_delineation_parameters.tex
\begin{table}[ht]
    \centering
    \begin{scriptsize}
    \renewcommand{\arraystretch}{1.3}
    \begin{tabular}{L{0.23\linewidth-2\tabcolsep}L{0.65\linewidth-2\tabcolsep}M{0.12\linewidth-2\tabcolsep}}
         Parameter & Description & Default Value  \\ \toprule
        Crown delineation voxel size & Voxel size for voxel-based down-sampling of the points before the crown delineation & \num{0.05}\,m \\
        $h_{seed}$ & Height of the cylindrical regions around the stems that are used for the selection of seed points & \num{0.6}\,m \\
        $f_{seed}$ & Factor to multiply with the stem diameter at breast height to obtain the diameter of the cylindrical regions around the stems that are used for the selection of seed points & \num{1.05} \\
        $\diameter_{min}^{seed}$ & Minimum diameter of the cylindrical regions around the stems that are used for the selection of seed points & \num{0.05}\,m \\
        $f_{z}$ & Factor by which to divide the z-coordinates of the points before the region growing (to promote upward growth) & \num{2} \\
        $r_{max}$ & Maximum region growing search radius & \num{0.5}\,m \\
        $t_{total}$ & Threshold controlling when to increase the region growing search radius. If in an iteration the ratio between the number of points newly assigned to trees and the number of remaining, unassigned points is below this threshold, the search radius is increased. & \num{0.2}\,\% \\
        $t_{trees}$ & Threshold controlling when to increase the region growing search radius. If in an iteration the ratio between the number of trees to which new points have been assigned and the total number of trees is below this threshold, the search radius is increased. & \num{30}\,\% \\
        $\Delta t_{\downarrow}$& Number of region growing iterations after which to decrease the search radius if it has not been increased for this number of iterations & \num{10}
        \\
        Max. region growing iterations & Maximum number of region growing iterations & \num{500} \\
        $d_{terrain}$ & Maximum cumulative search distance between an initial seed point and a terrain point to include that terrain point in a tree instance & \num{0.8}\,m
    \end{tabular}
    \end{scriptsize}
    \caption{Parameters of the tree crown delineation stage of our algorithm.}
    \label{tab:crown-segmentation-parameters}
\end{table}

%% file: sec_4_evaluation.tex
In our evaluation, we compared the accuracy and computational efficiency of our approach with those of six open-source tree instance segmentation approaches, including the original R implementation of the treeX algorithm provided by \citet{tockner-2022}. 
We used six publicly available datasets to evaluate our algorithm on a diverse set of point clouds covering different sensor types and forest characteristics.

\subsection{Datasets}

\paragraph{FOR-instance} The FOR-instance dataset~\citep{puliti-2023} is an aggregated dataset combining five subsets of \ac{ULS} point clouds from different institutions and countries (NIBIO, Norway; CULS, Czech 
Republic; TU~WIEN, Austria; RMIT, Australia; SCION; New Zealand). As such, the dataset covers forest stands of varying complexity and species composition, including coniferous-dominated temperate forests (CULS), coniferous-dominated 
boreal forests (NIBIO), deciduous-dominated alluvial forests (TU WIEN), dry-sclerophyll eucalypt forests (RMIT), and pure coniferous temperate forests (SCION). The data were acquired using UAV-mounted LiDAR sensors (Riegl VUX-1UAV and Riegl miniVUX-1UAV), with data acquisition protocols differing for the individual subsets. Except for the TU~WIEN subset, all data were acquired in leaf-on state. The authors of the dataset define a splitting scheme that partitions the data into development and test sets. Our evaluation was conducted using the test set, which consists of eleven point cloud files (\(1 \times \text{CULS}\), \(6 \times \text{NIBIO}\), \(1 \times \text{RMIT}\), \(2 \times \text{SCION}\), \(1 \times \text{TU~WIEN}\)).

\paragraph{ForestSemantic} The ForestSemantic dataset \citep{liang-2024} consists of \ac{TLS}~point clouds of six \num{32}\,m $\times$ \num{32}\,m boreal forest plots in the vicinity of Evo, Finland. However, only three of these point clouds are currently publicly available and were used in our evaluation (plot 1, plot 3, and plot 5). These plots are dominated by Scots pine (\textit{Pinus sylvestris L.}), Norway spruce (\textit{Picea abies L. Karst.}), silver birch (\textit{Betula pendula Roth}), and downy birch (\textit{Betula pubescens Ehrh.}). The data were captured in leaf-on state in April and May 2014 using a stationary \ac{TLS} scanner (Leica HDS6100), with five scans per plot (one in the center of the plot, four at the edges of the plot). The dataset represents a subset of a benchmarking dataset originally published by \cite{liang-2018}. In the original publication, the plots were assigned to three complexity classes depending on the density of the understory and the visibility of the stems (\enquote{easy}, \enquote{medium}, and \enquote{difficult}). The three publicly available plots of the ForestSemantic dataset cover all three complexity classes (plot~1 - easy, plot~3 - medium, plot~5 - difficult). The mean tree height of the plots decreases with increasing plot complexity, ranging from \num{18.7}\,m for plot~1 to \num{13}\,m for plot~5. However, the tree instance labels of the dataset only cover trees with a \ac{DBH} of more than \num{5}\,cm, and some understory trees are not labeled.

\paragraph{LAUTx} The LAUTx dataset~\citep{tockner-2022,gollob-2020} provides \ac{PLS}~point clouds of six circular forest plots located in Lower Austria, each covering \num{1257}\,m$^2$. The plots have diverse species compositions, with European beech (\textit{Fagus sylvatica L.}) and Norway spruce (\textit{Picea abies (L.) Karst.}) being the dominant species. Most of the plots are single-layered stands, and half of them have a moderate to dense understory. The data were collected in March 2019 in the leaf-off state using a handheld mobile laser scanner (GeoSLAM ZEB Horizon). The dataset published by \citet{tockner-2022} includes manually segmented point clouds corresponding to individual trees. \cite{henrich-2024} provided an updated set of labels for the dataset in which the tree instance labels are propagated to the complete plot-level point clouds, and some labeling errors are corrected, particularly in the segmentation of the tree bases. For our evaluation, we used the updated labels provided by \citet{henrich-2024}.

\paragraph{NIBIO~MLS} The NIBIO~MLS dataset \citep{wielgosz-2023} consists of \ac{PLS}~point clouds of several forest plots in south-eastern Norway. The plots cover areas with different species composition, with Norway spruce (\textit{Picea abies (L.) Karst.}), Scots pine (\textit{Pinus sylvestris L.}), or birch (\textit{Betula pubescens} or \textit{Betula pendula}) being the dominant species. The stands are mostly single-layered stands that are either in a mature stage or in the middle of their rotation period, with tree heights ranging from \num{15.6}\,m to \num{29}\,m.
The data were collected in June 2022, in leaf-on condition, using a mobile handheld laser scanner (GeoSLAM ZEB Horizon).
The dataset is partitioned into training, validation, and test sets. Our evaluation was conducted using the test set, which includes 3D~point clouds of 16 forest stands, each representing about one fourth of a circular forest plot of \num{2000}\,m$^2$.

\paragraph{TreeLearn} The TreeLearn dataset~\citep{henrich-2024} comprises \ac{PLS}~point clouds of 19 rectangular forest plots (\num{1.08} - \num{2.19}\,ha) distributed over four different regions of Germany, all dominated by European beech (\textit{Fagus sylvatica}). The data were captured in February 2021, in leaf-off condition, using a handheld mobile laser scanner (GeoSLAM ZEB Horizon) and subsampled to a \num{1}\,cm resolution. One of the 3D~point clouds in the dataset (L1W) was manually labeled, while the other 3D~point clouds were automatically labeled using the Lidar360 software. Since the labels obtained from the LiDAR360 software contain some errors, only the manually labeled 3D~point cloud was used in our evaluation. This 3D~point cloud represents a single-layered forest stand located near Lübeck, Germany, with an age of 136~years, an average tree height of \num{30}\,m, and moderate understory. Only trees with a height above \num{10}\,m are labeled in the dataset.

\paragraph{Wytham Woods} The Wytham Woods dataset~\citep{calders-2022,terryn-2020} consists of a \ac{TLS}~point cloud of a \num{1.52}\,ha forest plot near Oxford, UK, that is part of a larger study area run by Oxford University. The stand is dominated by European ash (\textit{Fraxinus excelsior}), Sycamore maple (\textit{Acer pseudoplatanus}), and English oak (\textit{Quercus robur}), with a dense understorey comprising of  Common hazel (\textit{Corylus avellana}) and Common hawthorn (\textit{Crataegus monogyna}). The plot contains both young and old-grown trees with heights ranging approximately from \num{1}\,m to \num{30}\,m. The dataset was acquired in leaf-off conditions, between late November 2015 and January 2016. A stationary terrestrial laser scanner (RIEGL VZ-400) was used for data collection, with scanning positions arranged in a grid of approximately \num{20}\,m $\times$ \num{20}\,m. The dataset was semi-automatically segmented into individual tree point clouds by manually correcting the results obtained from the treeseg software~\citep{burt-2019}. \citet{henrich-2024} created an updated set of labels for the dataset, in which the tree instance labels are refined and propagated to the complete plot-level point cloud. In this study, this updated version of the labels was used. Since the peripheral parts of the point cloud are unlabeled, we cropped a rectangular section from the central part of the point cloud. This cropped subset, which covers approximately \num{85}\,m \(\times\) \num{135}\,m was used for evaluation.
\\
\\
Our algorithm includes an optional stem filtering step based on reflection intensity values. However, the TreeLearn and Wytham Woods datasets do not include such values. Therefore, we processed these point clouds using only xyz-coordinates. For the other datasets, we used reflection intensity values as additional input. The ForestSemantic dataset's reflection intensity values range from \num{0} to \num{255}. We remapped these values to comply with the LAS specification,\footnote{LAS specification: \url{https://www.asprs.org/wp-content/uploads/2019/07/LAS_1_4_r15.pdf}} in which intensity values range from \num{0} to \num{65535}, consistent with the other datasets.

\subsection{Reference Approaches}

We compared our algorithm against six existing tree instance segmentation methods with publicly available open-source implementations. As a baseline, we included the original R~implementation of the treeX algorithm by \cite{tockner-2022}, as our method builds upon and revises their approach. As further representatives of unsupervised, algorithmic tree instance segmentation methods, we considered treeiso~\citep{xi-2022} and RayCloudTools~\citep{lowe-2021}. As representatives of recent deep learning methods, we included ForAINet~\citep{xiang-2024}, SegmentAnyTree~\citep{wielgosz-2024}, and TreeLearn~\citep{henrich-2024}. For these methods, we used the pre-trained model checkpoints provided by the respective authors and did not perform any additional training on the evaluation datasets. In the case of TreeLearn, multiple checkpoints are available corresponding to models trained on datasets of different sizes. We used the \enquote{small} variant, as the \enquote{mid} and \enquote{large} variants were trained on data that overlap with our evaluation datasets.

\subsection{Accuracy Metrics}

To evaluate the quality of our tree instance segmentation algorithm, we used established metrics that cover two quality aspects: Firstly, we considered instance detection metrics, which evaluate the results at the instance level by categorizing tree instances into \ac{TP}, \ac{FP}, and \ac{FN} instances. Secondly, we considered instance segmentation metrics that measure the point-wise segmentation quality for each reference instance, thereby reflecting how precisely the individual tree instances were delineated.

\subsubsection{Instance Detection Metrics}

To evaluate the quality of tree instance detection, the predicted instances must be matched with the reference instances to determine the number of true positives, false positives, and false negatives.
Several matching approaches have been proposed in the literature that calculate the degree of overlap between each predicted and each reference instance, using slightly different criteria to determine matches~\citep{vallet-2015,kirillov-2019,wielgosz-2023,henrich-2024}.
In accordance with~\cite{wielgosz-2024}, we adopted the matching approach proposed by \citet{kirillov-2019} since it provides an unambiguous matching between the predicted and reference instances. In this approach, the pairwise \ac{IoU} between each reference instance $\mathcal{R}_i$ and each predicted instance $\mathcal{P}_j$ is calculated:

\begin{equation}
     \text{IoU}(\mathcal{R}_i, \mathcal{P}_j) = \frac{| \mathcal{R}_i \cap \mathcal{P}_j |}{|\mathcal{R}_i \cup \mathcal{P}_j|} 
\end{equation}

If the IoU of $\mathcal{R}_i$ and $\mathcal{P}_j$ is strictly greater than $0.5$, both instances are matched. Predicted instances that match with a reference instance are counted as true positives, while predicted instances that do not match any reference instance are counted as false positives, and reference instances that do not match any predicted instance are counted as false negatives. Some of the test datasets used in our evaluation are not fully labeled. To avoid correct segmentations of unlabeled instances being counted as false positives, we only counted instances as false positives if more than \num{50}\,\% of the points assigned to a predicted instance were labeled. To prevent dense point cloud regions from dominating the matching, we down-sampled the point clouds using a \num{1}\,cm voxel grid before matching.
Following the evaluation protocol used by \cite{cherlet-2024}, we calculated the following instance detection metrics:
\begin{equation}
    \text{Precision} = \frac{\text{TP}}{\text{TP} + \text{FP}}
\end{equation}
\begin{equation}
    \text{Recall} = \frac{\text{TP}}{\text{TP} + \text{FN}}
\end{equation}
\begin{equation}
    \text{F$_1$-score} = \frac{2 \cdot \text{TP}}{2 \cdot \text{TP} + \text{FP} + \text{FN}} = 2 \cdot \frac{\text{Precision} \cdot \text{Recall}}{\text{Precision} +  \text{Recall}}
\end{equation}

\subsubsection{Instance Segmentation Metrics}

To evaluate the quality of point-wise tree instance segmentation, we followed the approach used by \citet{xiang-2024} and \citet{henrich-2024}. In this approach, the index \(\text{max}_{\mathcal{R}_i}\) of the predicted instance with the highest \ac{IoU} score is determined for each reference instance $\mathcal{R}_i$:

\begin{equation}
\text{max}_{\mathcal{R}_i} = \argmax \limits_{j=1, \dots, N_{\mathcal{P}}} (\text{IoU}(\mathcal{R}_i, \mathcal{P}_j))
\end{equation}

The \ac{mIoU} was then calculated by averaging the \ac{IoU} over all pairs of reference instances and the corresponding predicted instances:

\begin{equation}
\text{mIoU} = \frac{1}{N_\mathcal{R}} \sum_{i=0}^{N_\mathcal{R}} \text{IoU}(\mathcal{R}_i, \mathcal{P}_{max_{\mathcal{R}_i}})
\end{equation}

Note that, although this metric is referred to as \enquote{coverage} in the work of \cite{xiang-2024} and \cite{henrich-2024}, we use the term \enquote{\ac{mIoU}}, as it is more commonly used in the computer vision domain. In addition to the \ac{mIoU}, we computed \ac{mPrecision} and \ac{mRecall} as measures of commission and omission errors, respectively:

\begin{equation}
\text{mPrecision} = \frac{1}{N_\mathcal{R}} \sum_{i=0}^{N_\mathcal{R}} \frac{|\mathcal{R}_i \cap \mathcal{P}_{max_{\mathcal{R}_i}}|}{|\mathcal{P}_{max_{\mathcal{R}_i}}|}
\end{equation}

\begin{equation}
\text{mRecall} = \frac{1}{N_\mathcal{R}} \sum_{i=0}^{N_\mathcal{R}} \frac{|\mathcal{R}_i \cap \mathcal{P}_{max_{\mathcal{R}_i}}|}{|\mathcal{R}_i|}
\end{equation}

As in the computation of instance detection metrics, we down-sampled the point clouds with a \num{1}\,cm voxel grid before the metric computation.

\subsubsection{Metric Aggregation}

For test datasets comprising multiple 3D~point cloud files, we aggregated the metrics at the instance level. For instance detection, we summed the numbers of true positives, false positives, and false negatives across all point clouds, and then computed precision, recall, and F$_1$-score. For instance segmentation, we gathered all matched pairs of predicted and reference instances from the individual 3D~point clouds and computed the average metrics over these pairs. This approach weights each 3D~point cloud according to the number of trees it contains.

\subsection{Computational Efficiency Metrics}

To evaluate the computational efficiency of our algorithm, we measured wall-clock time, CPU time, and peak memory usage.
Wall-clock time refers to the total elapsed real-world time from the start to the end of the algorithm’s execution. It includes both active computation and any waiting time (including disk I/O) and can therefore be affected by other processes running concurrently on the system.
In contrast, CPU time only measures the time during which the algorithm actively runs on the CPU.
In the case of parallel execution across multiple CPU cores, CPU time accumulates across all cores, providing a measure of an algorithm's total computational effort.
Peak memory usage is captured using the maximum resident set size, which represents the maximum amount of physical memory allocated to the process during execution, excluding memory pages swapped to disk.
We used the \textit{perf\_counter} of Python's time module to measure wall-clock time. CPU time and peak memory usage were measured using the resource package (CPU time: \textit{ru\_utime} + \textit{ru\_stime}, peak memory usage: \textit{ru\_maxrss}).

\subsection{Hardware and Software Setup}

All tested tree instance segmentation methods were containerized using Docker and executed on a SLURM-managed high-performance computing cluster under consistent conditions. For each combination of segmentation method and 3D~point cloud file, a separate SLURM array task was launched.

As some of the tested tree instance segmentation methods require substantial computational resources, the main evaluation experiments were conducted using a high-end hardware configuration. Each SLURM task was allocated 32~CPU cores of an AMD EPYC 7742 processor (\num{2.25}\,GHz) and an NVIDIA A100 GPU with \num{40}\,GB VRAM. The amount of main memory allocated per task was adapted to the dataset size: By default, \num{64}\,GB were allocated for the FOR-instance and NIBIO MLS datasets, \num{128}\,GB for the ForestSemantic, LAUTx, and TreeLearn datasets, and \num{256}\,GBfor the Wytham Woods dataset. In cases of out-of-memory errors, the memory budget for the affected methods was increased until they ran successfully.

The ablation studies were conducted with reduced computational resources. Specifically, each SLURM task was allocated 16 cores of an AMD EPYC 7742 CPU (\num{2.25}\,GHz) and \num{32}\,GB main memory. No GPU was used in this configuration.

%% file: sec_5_results.tex
We first analyze the accuracy of our algorithm and compare it with existing open-source tree instance segmentation methods. Next, we compare the methods in terms of their computational efficiency. Finally, we present ablation studies in which individual components of our algorithm were replaced to examine their impact on overall performance.

\input{fig_example_results}

\subsection{Instance Detection and Segmentation Accuracy}

\input{fig_instance_detection_metrics}
\input{fig_instance_segmentation_metrics}
\input{fig_segmentation_errors}

We begin by comparing the accuracy of our version of the treeX algorithm with that of the original implementation and evaluating our presets. In the following, we compare the accuracy with that of other open-source tree instance segmentation approaches. The evaluation metrics are shown in Fig.~\ref{fig:instance-detection-metrics} and Fig.~\ref{fig:instance-segmentation-metrics}.

\paragraph{Comparison with the original treeX algorithm} The results show that our revised version of the treeX algorithm substantially improves the accuracy of the original version. The revised version achieved higher detection and segmentation metrics on all test datasets, except for the ForestSemantic dataset, where the original algorithm achieved slightly higher instance detection precision.

\paragraph{Comparison of our parameter presets} The parameter presets introduced by us proved to be beneficial: While the \ac{TLS} preset failed to produce any true positives on the \ac{ULS} point clouds of the FOR-instance dataset, the \ac{ULS} preset achieved a detection F$_1$-score of \num{0.58} and a segmentation \ac{mIoU} of \num{0.47}. Conversely, the \ac{TLS} preset consistently outperformed the \ac{ULS} preset on the \ac{TLS} and \ac{PLS} datasets.

\paragraph{Comparison with reference methods for ULS data} Despite the improvements introduced by our \ac{ULS} preset, our algorithm remains better suited for \ac{TLS} and {PLS} data with high point densities in the stem layer. Consequently, on \ac{ULS} data, our algorithm was outperformed by all deep learning approaches. The tested algorithmic approaches, RayCloudTools and treeiso, produced lower or similar detection F$_1$-scores as our \ac{ULS} preset but higher segmentation \ac{mIoU} scores.

\paragraph{Comparison with reference methods for TLS and PLS data} Considering the accuracy of our \ac{TLS} preset on \ac{TLS} and \ac{PLS} data shows that our method is competitive with recent deep learning methods for this type of data. Our algorithm outperformed SegmentAnyTree and ForAINet on the ForestSemantic, LAUTx, TreeLearn, and Wytham Woods datasets, while the two deep learning methods achieved better results on NIBIO MLS. Conversely, TreeLearn performed better than our approach on the TreeLearn and Wytham Woods datasets, but was outperformed by our method on the NIBIO MLS and ForestSemantic datasets. For the LAUTx dataset, our algorithm and TreeLearn performed similarly, with our algorithm achieving a higher detection F$_1$-score and TreeLearn achieving a higher segmentation \ac{mIoU}.

Compared to RayCloudTools, a well-performing algorithmic approach~\citep{cherlet-2024}, our algorithm also demonstrates competitive performance on \ac{TLS} and \ac{PLS} data. 
On the LAUTx, NIBIO MLS, and Wytham Woods datasets, the performance of both methods was comparable in terms of detection F$_1$-scores. In general, RayCloudTools tends to produce fewer omission errors, resulting in higher detection recall, whereas our algorithm makes fewer commission errors, leading to higher detection precision.
On the ForestSemantic and TreeLearn datasets, our algorithm outperformed RayCloudTools. For the TreeLearn dataset, the instance detection precision of RayCloudTools was particularly low, primarily due to its detection of a large number of understory trees. Although some of these detections may be correct, they were counted as false positives because smaller understory trees are not labeled in the TreeLearn dataset. For the ForestSemantic dataset, RayCloudTools performed poorly on the medium and difficult plots, leading to both low instance detection precision and recall.

Treeiso, another representative of algorithmic approaches, was outperformed by our algorithm on all datasets.

\paragraph{Error analysis of our algorithm} By visually inspecting the segmentation results of our algorithm, we identified common types of error. Examples of typical errors are shown in Fig.~\ref{fig:segmentation-errors}, and error visualizations for all evaluation datasets are provided in the supplementary material. Our algorithm relies on detecting dense clusters of stem points within a certain height range above the ground (\num{1}\,m to \num{4}\,m for the \ac{TLS} preset, \num{1}\,m to \num{5}\,m for the \ac{ULS} preset). Most segmentation errors are related to this stem detection stage. For example, trees whose stems are largely obscured by dense branches or understory vegetation within the relevant height range were often not detected (Fig.~\ref{fig:errors-nibio-mls}, Fig.~\ref{fig:errors-forest-semantic}).
In addition, stems with insufficient point density were missed in several cases. This issue occurred mainly in the FOR-instance dataset (Fig.~\ref{fig:errors-for-instance}), which has a low point density in the stem layer, which is characteristic of \ac{ULS} point clouds, as well as in the sparse peripheral regions of the ForestSemantic and NIBIO MLS datasets. Notably, no trees were detected in the \textit{TU Wien} point cloud from the FOR-instance dataset and the \textit{burum2} point cloud from the NIBIO MLS dataset, both of which are very sparse.
In such cases of omission errors, where some stems are not detected, the corresponding trees are often merged into neighboring tree instances (Fig.~\ref{fig:errors-nibio-mls}, Fig.~\ref{fig:errors-forest-semantic}).
Although this case did not occur in the evaluation datasets, omission errors may also occur for strongly inclined stems, as they form less dense clusters in the 2D~clustering step of our algorithm.

Conversely, the stem detection stage occasionally produces false positives in cases where dense point clusters are formed by parts of the tree crown. This issue particularly occurred in coniferous stands included in the FOR-instance and NIBIO MLS datasets, where dense crowns with downward-facing branches fall within the stem detection height range.
In the Wytham Woods dataset, we observed another source of false positives: Stems were frequently fragmented into two point clusters (Fig.~\ref{fig:errors-wytham-woods}). This is likely because the dataset was generated from multiple \ac{TLS} scans taken at different positions. This results in uneven point densities along the stems, with high-density areas separated by lower-density gaps.
Such false positive stem detections lead to oversegmentation, where parts of the stems or crowns are incorrectly split into separate instances.

Apart from errors related to stem detection, our algorithm also shows some inaccuracies in delineating very dense and interwoven tree crowns. These issues are caused by the region growing procedure used for crown delineation and mainly occurred in dense coniferous stands with relatively low point densities in the FOR-instance dataset (Fig.~\ref{fig:errors-for-instance}).

We also observed a few errors related to issues with the reference labels rather than actual errors made by our algorithm. In the ForestSemantic and the TreeLearn datasets, many understory trees are not labeled. For plots 1 and 3 of the ForestSemantic dataset, our algorithm correctly segmented several understory trees, which were then incorrectly counted as false positives (Fig.~\ref{fig:errors-forest-semantic}).
In the NIBIO MLS dataset, the test plots are subsections of larger scans and therefore often include tree instances in peripheral regions that consist of only a few crown points. These instances were commonly not detected by our algorithm, resulting in a low recall (Fig.~\ref{fig:errors-nibio-mls}).

Consistent with the quantitative metrics, visual inspection confirmed the high accuracy of our algorithm for the LAUTx and TreeLearn datasets (Fig.~\ref{fig:example-results-tree-learn}). In these datasets, most of the overstory trees were segmented accurately, with only minor errors in cases where understory vegetation was close to the stems.

\input{fig_computational_resources_detailed}
\input{fig_computational_resources}
\input{fig_ablation_study_tls}

\subsection{Computational Efficiency}

In Fig.~\ref{fig:computational-resources}, the computational performance of our algorithm is compared with existing tree instance segmentation methods using representative point clouds from the evaluation datasets. For the FOR-instance and NIBIO MLS examples, the processing times were low for all tested methods due to the small point cloud sizes (wall-clock times below \num{5}\,min for all methods). For the larger point clouds from the other datasets, the wall-clock time of our algorithm (values for \ac{TLS} preset) ranged from \num{21}\,min for plot~1 of the ForestSemantic dataset (\(\approx\) \num{205}\,min/ha) to \num{140}\,min for Wytham Woods (\(\approx\) \num{122}\,min/ha). Compared to the original version of the treeX algorithm, our version achieved \num{1.6} to \num{2.8} times faster wall-clock times. However, the CPU time, which reflects the total number of CPU operations executed, was up to \num{2.2} times higher than that of the original version. This indicates that the faster wall-clock time of the revised version is primarily due to the parallelization of computations across multiple CPU cores. The runtime of the original version could potentially be improved with similar parallelization strategies.
The ratio of CPU time to wall-clock time of our version is substantially lower than the number of CPU cores used (max. ratio of \num{1.7} for 32 CPU cores). This is because only certain steps of our algorithm, such as circle fitting and neighbor searches during region growing, are implemented in a parallelized manner. Consequently, processing a single point cloud does not fully utilize all available CPU resources. To do so, multiple point cloud files could be processed simultaneously.

Fig.~\ref{fig:runtime-detailed} presents a detailed breakdown of the runtime across the individual steps of our algorithm. The region growing step is by far the most computationally demanding, accounting for up to \num{87}\,\% of the total runtime in the case of the Wytham Woods dataset. We found that down-sampling the point clouds before region growing provides a trade-off between runtime and accuracy. Stronger down-sampling reduces execution time at the cost of lower segmentation accuracy. In our evaluations, we used a default downsampling voxel size of \num{0.05}\,m to balance runtime and accuracy (Table~\ref{tab:crown-segmentation-parameters}).

Comparing runtimes with other tree instance segmentation approaches shows that RayCloudTools, treeiso, and TreeLearn achieve faster execution times than our algorithm (Fig.~\ref{fig:computational-resources}). In particular, RayCloudTools demonstrates high computational efficiency. In terms of wall-clock time, it outperformed our algorithm by a factor of \num{1.7} to \num{4.5} across all point cloud files, except for plot~1 from the ForestSemantic dataset. For example, processing the Wytham Woods dataset took \num{42}\,min with RayCloudTools compared to \num{140}\,min with our algorithm. However, for the ForestSemantic example, RayCloudTools had a \num{6.2} times longer wall-clock time than our algorithm (\num{126}\,min vs. \num{21}\,min). TreeLearn achieved fast execution times across all datasets, but benefits from utilizing additional GPU resources. Despite also using GPU acceleration, SegmentAnyTree and ForAINet had substantially slower runtimes than our algorithm.

In terms of main memory consumption, our algorithm demonstrates efficient performance. For the larger sample point clouds from the ForestSemantic, LAUTx, TreeLearn, and Wytham Woods datasets, all tested methods, except for the original version of the treeX algorithm, exhibited higher memory consumption. Our measurements only considered main memory and did not include GPU memory. Unlike deep learning methods, our algorithm does not utilize GPU memory. For Wytham Woods, which is the largest point cloud in our evaluation (\num{201}\,M points), the \ac{TLS} preset of our algorithm required \num{45}\,GB of main memory for processing. The \ac{ULS} preset consumed more memory due to different DBSCAN clustering settings that produce larger clusters. However, this is not a practical concern, as the \ac{ULS} preset is specifically designed for sparser \ac{ULS} point clouds. Among the compared methods, treeiso showed particularly high memory consumption, requiring \num{503}\,GB for processing the TreeLearn dataset and \num{871}\,GB for Wytham Woods.

\subsection{Ablation Studies}

In our revised version of the treeX algorithm, we have modified, removed, and added steps compared to the original version. To assess the impact of these changes on both accuracy and computational performance, we conducted two ablation studies: The first ablation study evaluates the effectiveness of the modifications introduced in our \ac{TLS} preset compared to the original version of the treeX algorithm. The second ablation study examines the effectiveness of the adjustments made in our \ac{ULS} preset relative to our \ac{TLS} preset.

\subsection{Ablation Study for the TLS preset}

In the first ablation study, we used our \ac{TLS} preset as the baseline and tested several modifications that reflect settings used in the original version of the treeX algorithm. All tested modifications are fully integrated into the algorithm’s implementation and can be activated via parameter settings on demand. Since the \ac{TLS} preset is designed for dense \ac{TLS} and \ac{PLS} point clouds, the experiments of the first ablation study were limited to the \ac{TLS} and \ac{PLS} datasets (ForestSemantic, LAUTx, NIBIO MLS, TreeLearn, Wytham Woods). As shown in Fig.~\ref{fig:ablation-study-tls}, the following modifications were evaluated:

\paragraph{Layer configuration} The stem detection stage of the treeX algorithm involves fitting circles to multiple horizontal layers extracted from the potential stem clusters. As detailed in Table~\ref{tab:stem-detection-parameters}, the position and extent of these layers is controlled by the parameters \(h_{\min}\) (lower boundary of the stem layer used to identify the stem clusters), \(h_{\max}\) (upper boundary of the stem layer used to identify the stem clusters), \(N_l\) (number of horizontal layers extracted from each stem cluster), \(h_0\) (height of the first extracted layer above the ground), \(h_l\) (height of the extracted layers), and \(o_l\) (vertical overlap between adjacent layers). In this experiment, we set these parameters to the values used in the original implementation of the treeX algorithm~\citep{gollob-2020}: \(h_{\min} = \)~\num{1}\,m, \(h_{\max} = \)~\num{3}\,m, \(N_l = 14\), \(h_0 = \)~\num{1}\,m, \(h_l = \)~\num{0.15}\,m, \(o_l = \)~\num{0.025}\,m. Compared to these original settings, our \ac{TLS} preset achieved higher instance detection F\(_1\)-scores for all test datasets, with particularly notable increases for the TreeLearn and Wytham Woods datasets (\num{+0.09} each). The impact on computational performance was mixed. In particular, the CPU time for TreeLearn (\num{-16}\,min) and Wytham Woods (\num{-29}\,min) was decreased by our preset, but the memory consumption for plot~1 of the LAUTx dataset (\num{+4.1}\,GB) was increased.

\paragraph{Circle fitting} In this experiment, we replaced the RANSAC circle fitting algorithm in the stem detection stage of our implementation with the gradient-based circle fitting method from the original version of the treeX algorithm.
Compared to the original implementation, our modification scored notably higher in instance detection F\(_1\)-scores on all tested datasets apart from ForestSemantic (\num{-0.01}). Particularly noticeable is the increase in accuracy for TreeLearn (\num{+0.06}) and Wytham Woods (\num{+0.07}). Additionally,  the RANSAC approach is faster and less memory-demanding compared to the gradient-based circle fitting of the original implementation. 

\paragraph{+Refined circle fitting} The original version of the treeX algorithm employs a two-stage circle fitting approach~\citep{gollob-2020}. In the first stage, circles are fitted to the potential stem clusters, using a down-sampled version of the point cloud. In the second stage, all points within a buffer around the outline of the initially fitted circles are extracted from the full-resolution point cloud and used to refine the circle fitting. By default, our revised version of the treeX algorithm only includes the first circle fitting stage. In this experiment, we additionally incorporated the second stage for refined circle fitting. This resulted in slightly improved instance detection F\(_1\)-scores for the ForestSemantic dataset (\num{+0.02}) and the TreeLearn dataset (\num{+0.01}). However, it also significantly increased CPU time (\num{+19}\,min for ForestSemantic, plot 1, and \num{+27}\,min for TreeLearn). Given the high computational cost, we chose not to include the refined circle fitting in the default settings of our algorithm. Nevertheless, since our implementation parallelizes the circle fitting across multiple CPU cores, the impact on wall-clock time remains low when sufficient CPU resources are available. In such settings, it is recommended to enable the refined circle fitting to further improve segmentation accuracy.

\paragraph{+Ellipse fitting} The original version of the treeX algorithm uses ellipse fitting in addition to circle fitting, because circle fitting may be unsuccessful in some cases~\citep{gollob-2020}. In this experiment, we included ellipse fitting in our algorithm. This resulted in slight improvements in instance detection F\(_1\)-scores for the ForestSemantic, LAUTx, NIBIO MLS datasets (\num{+0.01} each), but led to a decrease for the TreeLearn dataset (-\num{0.04}). The computational cost of the ellipse fitting step was low for most test datasets. However, since we did not observe a consistently positive impact on detection F\(_1\)-scores, we disabled ellipse fitting in the default settings of our algorithm.

\paragraph{{\textminus}Circular completeness index filter} To filter out invalid circles, we newly introduced a filtering step based on the \ac{CCI}~\citep{krisanski-2020} in our revised version of the treeX algorithm.
Compared to a setting without \ac{CCI}-based filtering, the F\(_1\)-score improved for TreeLearn (\num{+0.08}) and Wytham Woods (\num{+0.01}). For the NIBIO MLS dataset,  the F\(_1\)-score dropped slightly (\num{-0.01}), whereas for TreeLearn and Wytham Woods, slightly higher CPU time (\num{+4}\,min and \num{+3}\,min) and memory consumption (\num{+0.2}\,GB each) were observed.

\paragraph{Circle diameter standard deviation filter} After fitting circles to different layers of the potential stem clusters, the standard deviation of the circle diameters is computed for different subsets of the layers and stem clusters with too high standard deviation are filtered out. In our revised version of the treeX algorithm, we increased the filtering threshold \(\sigma^{\diameter}_{\max}\) to \num{0.04}\,m compared to the original threshold of \num{0.0185}\,m~\citep{gollob-2020}.
This modification led to an improvement of the instance detection F\(_1\)-score for all tested datasets (\num{+0.01} for ForestSemantic and Wytham Woods, \num{+0.03} for LAUTx, and \num{+0.04} for NIBIO MLS) except for TreeLearn (\num{-0.01}). Regarding the computational costs, the CPU time decreased for LAUTx (\num{-3}\,min), TreeLearn (\num{-20}\,min), and WythamWoods (\num{-6}\,min) datasets, while slightly increasing for the Forest\-Semantic (\num{+2}\,min) dataset.

\paragraph{+PCA explained variance filter} During the development of the revised treeX algorithm, we experimented with an additional filtering criterion for potential stem clusters. This filter applies principal component analysis to the cluster points and discards clusters where the variance explained by the first principal component is below a specified threshold. In this experiment, we set the threshold to \num{0.5}. Since this modification did not change the instance detection F\(_1\)-score, we disabled this filter by default.

\paragraph{+PCA inclination filter} Using principal component analysis as in the filter described before, we also tested a filter that discards potential stem clusters if their inclination (angle between the first principal component and the z-axis) exceeds a certain threshold. In this experiment, we set this threshold to \num{45}~\textdegree. Since this did not impact the instance detection F\(_1\)-score, we also disabled this filter by default.

\input{fig_ablation_study_uls}

\subsection{Ablation Study for the ULS preset}

In the second ablation study, we used the \ac{ULS} preset as the baseline and tested several modifications that reflect settings used in our \ac{TLS} preset. Since the \ac{ULS} preset is designed for sparse \ac{ULS} point clouds, the experiments of the ablation study were limited to the FOR-instance dataset. The following modifications were evaluated (Fig.~\ref{fig:ablation-study-uls}):

\paragraph{DBSCAN clustering} To account for the point sparsity in \ac{ULS} point clouds, we adjusted the parameters of the 2D and 3D~DBSCAN clustering steps for the \ac{ULS} preset. The \ac{ULS} preset achieved an instance detection F\(_1\)-score of \num{0.57} on the FOR-instance dataset. In contrast, with the clustering settings of the \ac{TLS} preset, no stems were detected at all, resulting in a F\(_1\)-score of zero. Since the algorithm terminates in this case, the computational efficiency was not comparable.

\paragraph{Minimum number of points per cluster} The \ac{ULS} preset uses a reduced value for the filtering threshold \(N_{\min}\) regarding the minimum number of points per stem cluster to accommodate point sparsity (\(N_{min} = 20\)). Compared to the \ac{TLS} preset value of \(N_{\min} = 300\), this increased the detection F\(_1\)-score by \num{0.11}.
 
\paragraph{Layer configuration} The stem detection stage of the treeX algorithm involves fitting circles to multiple horizontal layers extracted from potential stem clusters. Compared to the \ac{TLS} preset, the \ac{ULS} preset uses layers with a larger vertical extent to ensure that each layer contains a sufficient number of points for circle fitting. Additionally, the \ac{ULS} preset reduces the number of layers sampled for filtering based on the standard deviation of circle diameters. Compared to the corresponding settings of the \ac{TLS} preset (\(h_{\min} = \)~\num{1}\,m, \(h_{max} = \)~\num{4}\,m, \(h_0 = \)~\num{1}\,m, \(N_l = 15\), \(h_l = \)~\num{0.225}\,m, \(o_l = \)~\num{0.025}\,m, \(N_{sample} = 6\)), the modified \ac{ULS} setting led to a substantial increase of the instance detection F\(_1\)-score (\num{+0.22}).

\paragraph{Circle fitting} In this experiment, the circle fitting settings of the \ac{ULS} preset were compared to the corresponding settings of the \ac{TLS} preset (circle fitting bandwidth = \(0.01\), minimum number of points required for circle fitting = \(15\), minimum fitting score = \(100\)). The circle fitting settings of the \ac{ULS} preset improved the instance detection F\(_1\)-score by \num{0.11}.

\paragraph{Circle diameter standard deviation filter} After fitting circles to different layers of the potential stem clusters, the standard deviation of the circle diameters is computed for various subsets of the layers, and stem clusters with too high standard deviation are filtered out. Since \ac{ULS} point clouds typically exhibit higher noise levels than \ac{TLS} and \ac{PLS} point clouds, the threshold for the standard deviation was increased in the \ac{ULS} preset. Compared to the \ac{TLS} preset threshold value of \(\sigma^{\diameter}_{\max} = 0.04\), this slightly increased the instance detection F\(_1\)-score by \num{0.02}.

%% file: fig_example_results.tex
\begin{figure}
    \centering
    \includegraphics[width=\linewidth]{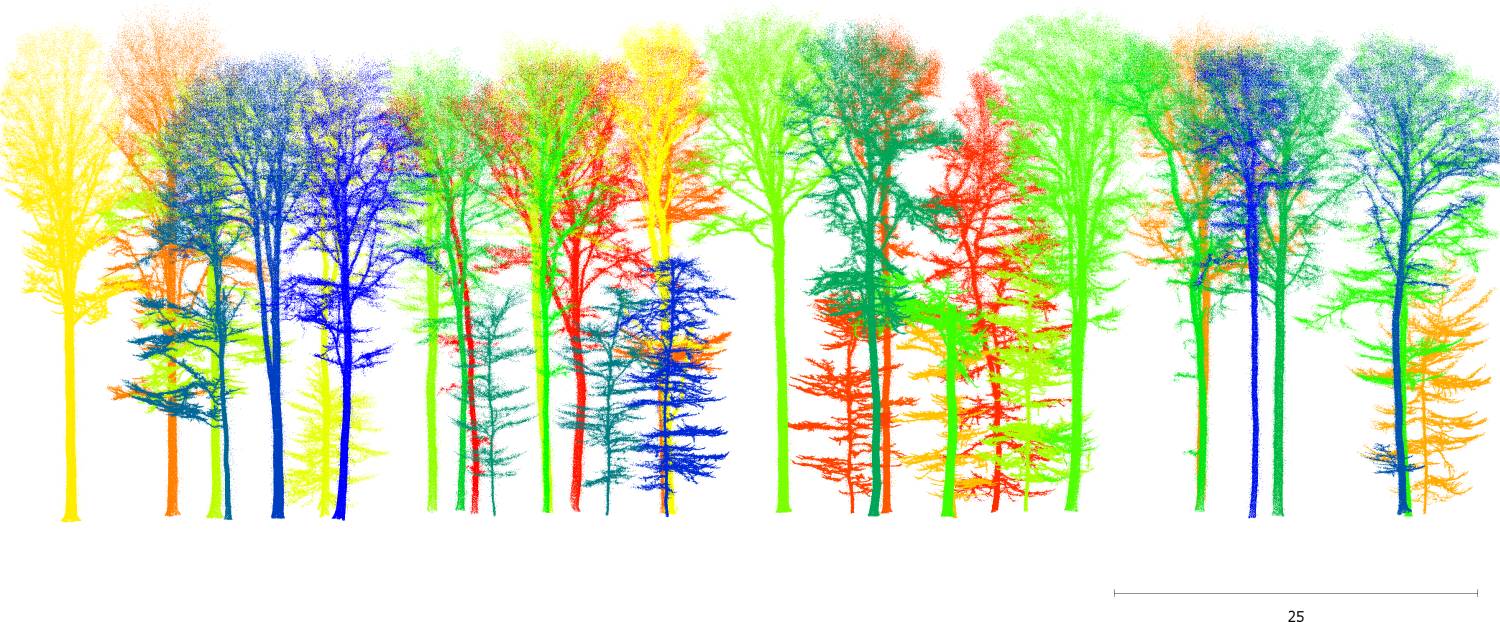}
    \caption{Example result of our algorithm (TLS preset) for a subsection of the TreeLearn (L1W) dataset (different colors represent different tree instances).}
    \label{fig:example-results-tree-learn}
\end{figure}

%% file: fig_instance_detection_metrics.tex
\begin{figure*}
\centering
\begin{scriptsize}
\begin{subfigure}[c]{\linewidth}
\centering
\includegraphics{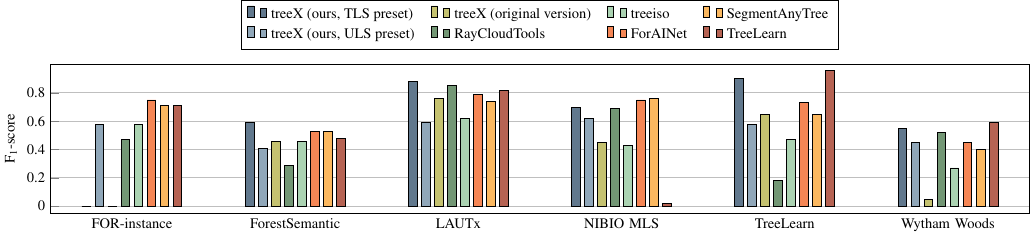}
\vspace{-0.2cm}
\caption{F\(_1\)-score}
\label{fig:instance-detection-f1-score}
\vspace{0.2cm}
\end{subfigure}
\begin{subfigure}[c]{\linewidth}
\centering
\includegraphics{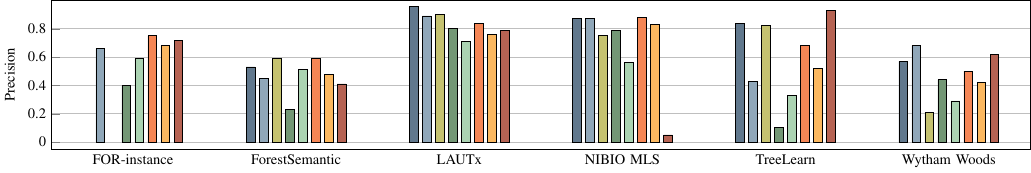}
\vspace{-0.2cm}
\caption{Precision}
\label{fig:instance-detection-precision}
\vspace{0.2cm}
\end{subfigure}
\begin{subfigure}[c]{\linewidth}
\centering
\includegraphics{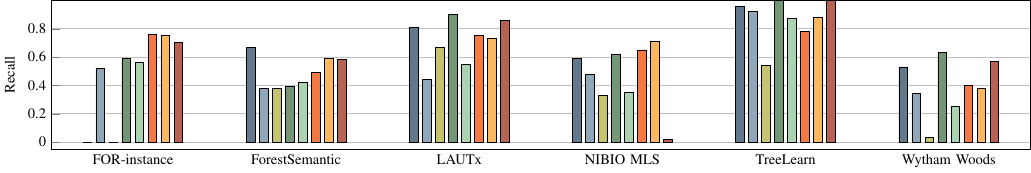}
\vspace{-0.2cm}
\caption{Recall}
\label{fig:instance-detection-recall}
\end{subfigure}
\caption{Comparison of the instance detection metrics of our algorithm with existing open-source tree instance segmentation methods.}
\label{fig:instance-detection-metrics}
\end{scriptsize}
\end{figure*}

%% file: fig_instance_segmentation_metrics.tex
\begin{figure*}
\centering
\begin{scriptsize}
\begin{subfigure}[c]{\linewidth}
\centering
\includegraphics{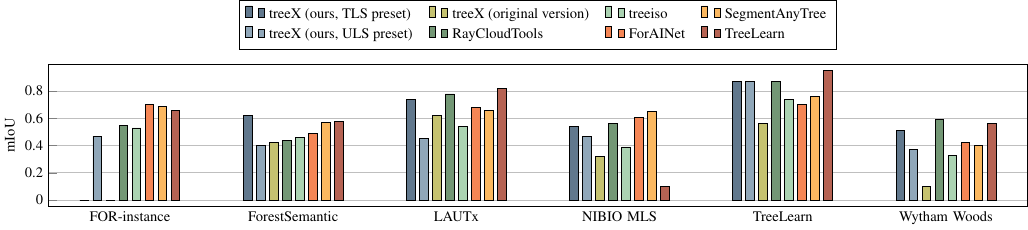}
\vspace{-0.2cm}
\caption{Mean IoU}
\label{fig:instance-segmentation-f1-score}
\vspace{0.2cm}
\end{subfigure}
\begin{subfigure}[c]{\linewidth}
\centering
\includegraphics{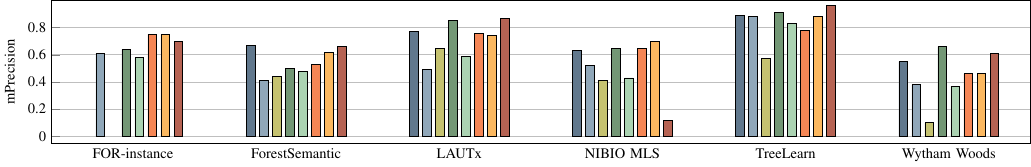}
\vspace{-0.2cm}
\caption{Mean precision}
\label{fig:instance-segmentation-precision}
\vspace{0.2cm}
\end{subfigure}
\begin{subfigure}[c]{\linewidth}
\centering
\includegraphics{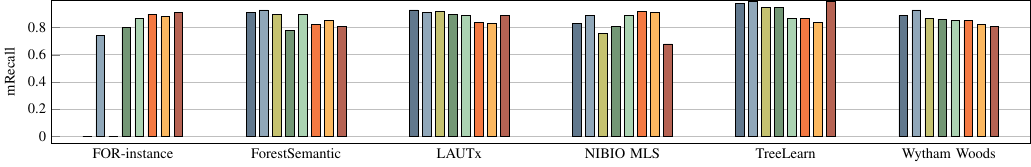}
\vspace{-0.2cm}
\caption{Mean recall}
\label{fig:instance-segmentation-recall}
\end{subfigure}
\caption{Comparison of the instance segmentation metrics of our algorithm with existing open-source tree instance segmentation methods.}
\label{fig:instance-segmentation-metrics}
\end{scriptsize}
\end{figure*}

%% file: fig_segmentation_errors.tex
\begin{figure*}
    \centering
    %
    %
    %
    \begin{subfigure}[t]{0.48\linewidth}
    \begin{subfigure}[t]{0.49\linewidth}
    \centering
    \includegraphics[width=\linewidth]{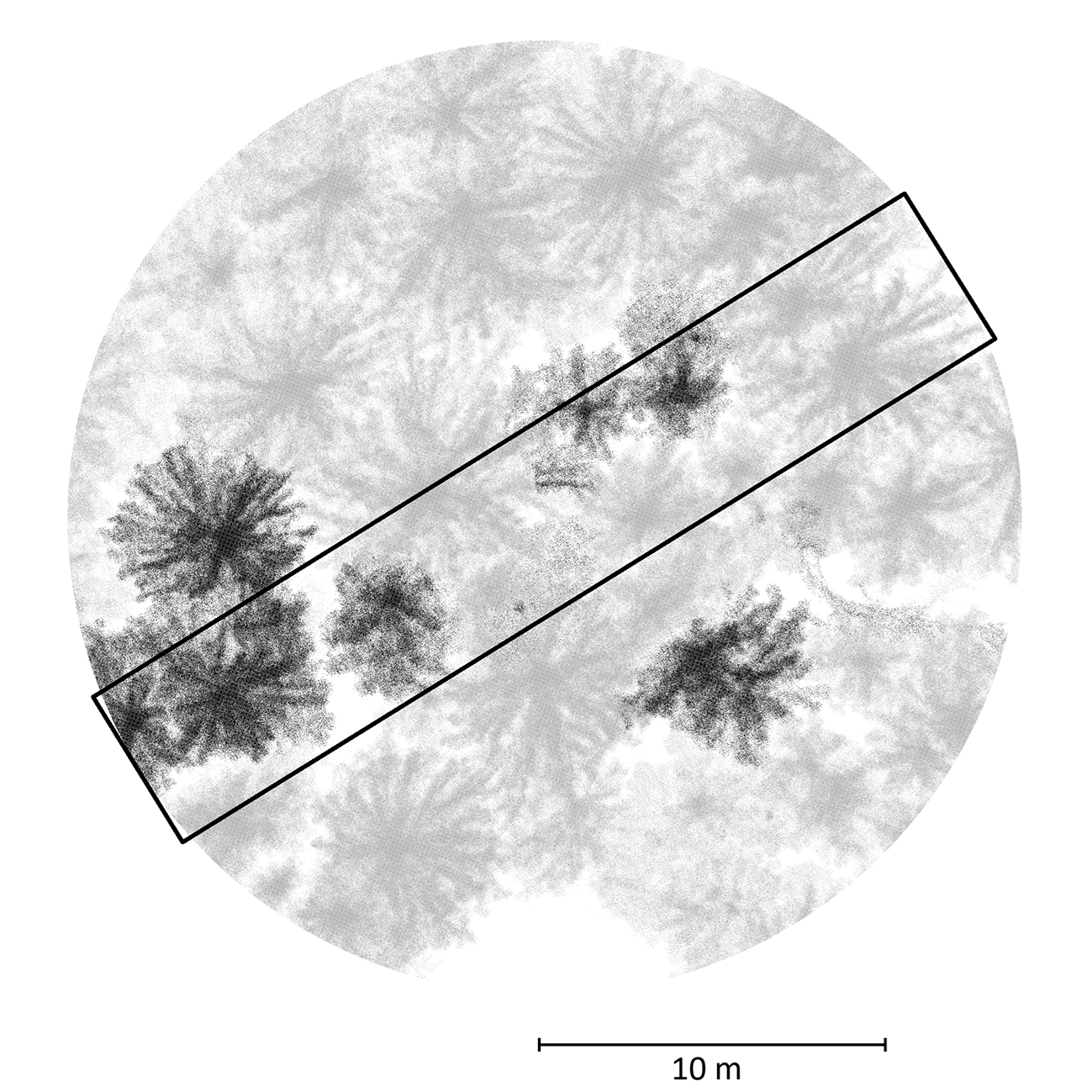}
    \end{subfigure}
    \hfill
    \begin{subfigure}[t]{0.49\linewidth}
    \centering
    \includegraphics[trim={200px 0 0 0}, clip, width=\linewidth]{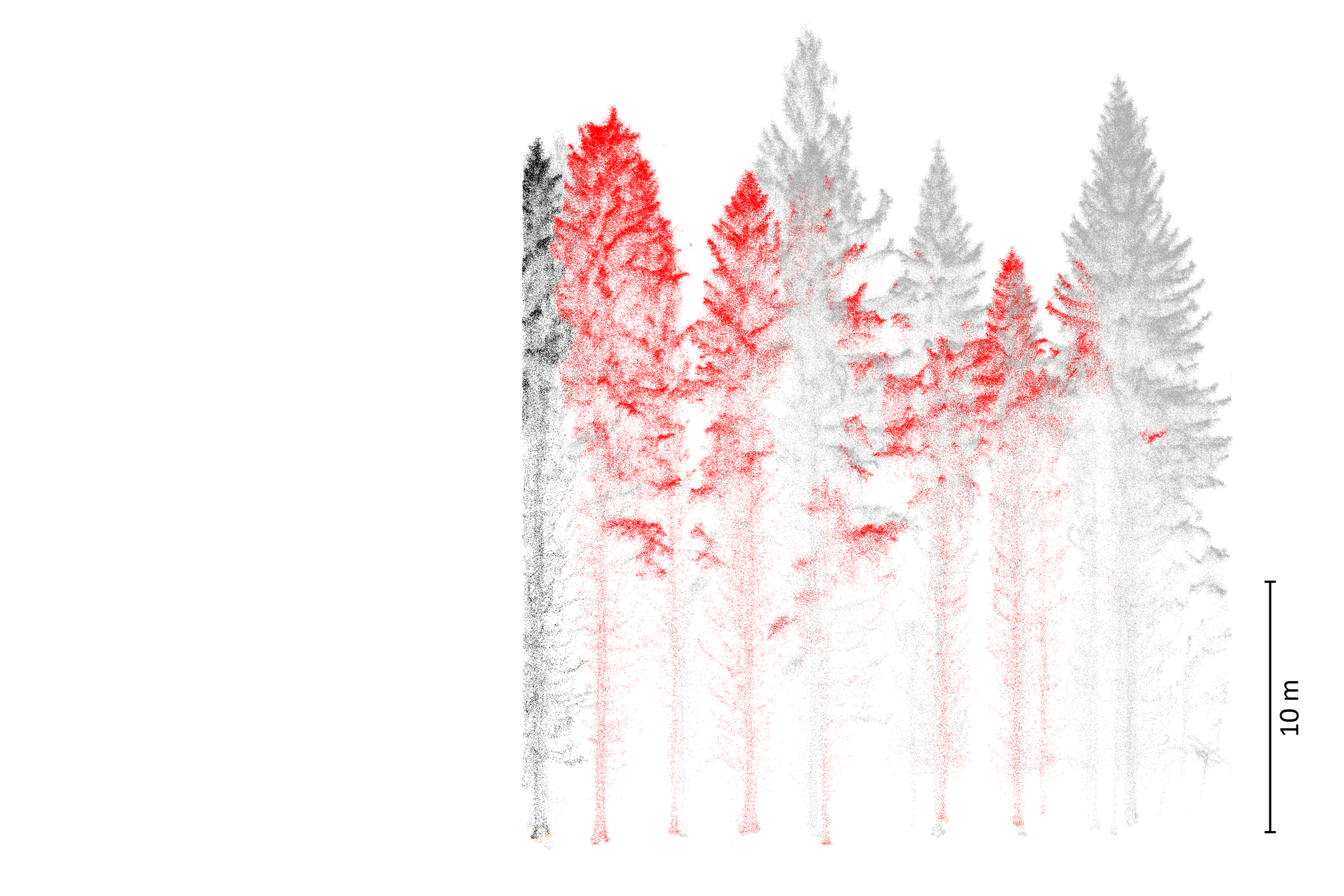}
    \end{subfigure}
    \caption{FOR-instance, NIBIO, plot 1}
    \label{fig:errors-for-instance}    \end{subfigure}
    %
    %
    %
    \hfill
    \begin{subfigure}[t]{0.45\linewidth}
    \begin{subfigure}[t]{0.49\linewidth}
    \centering
    \includegraphics[width=\linewidth]{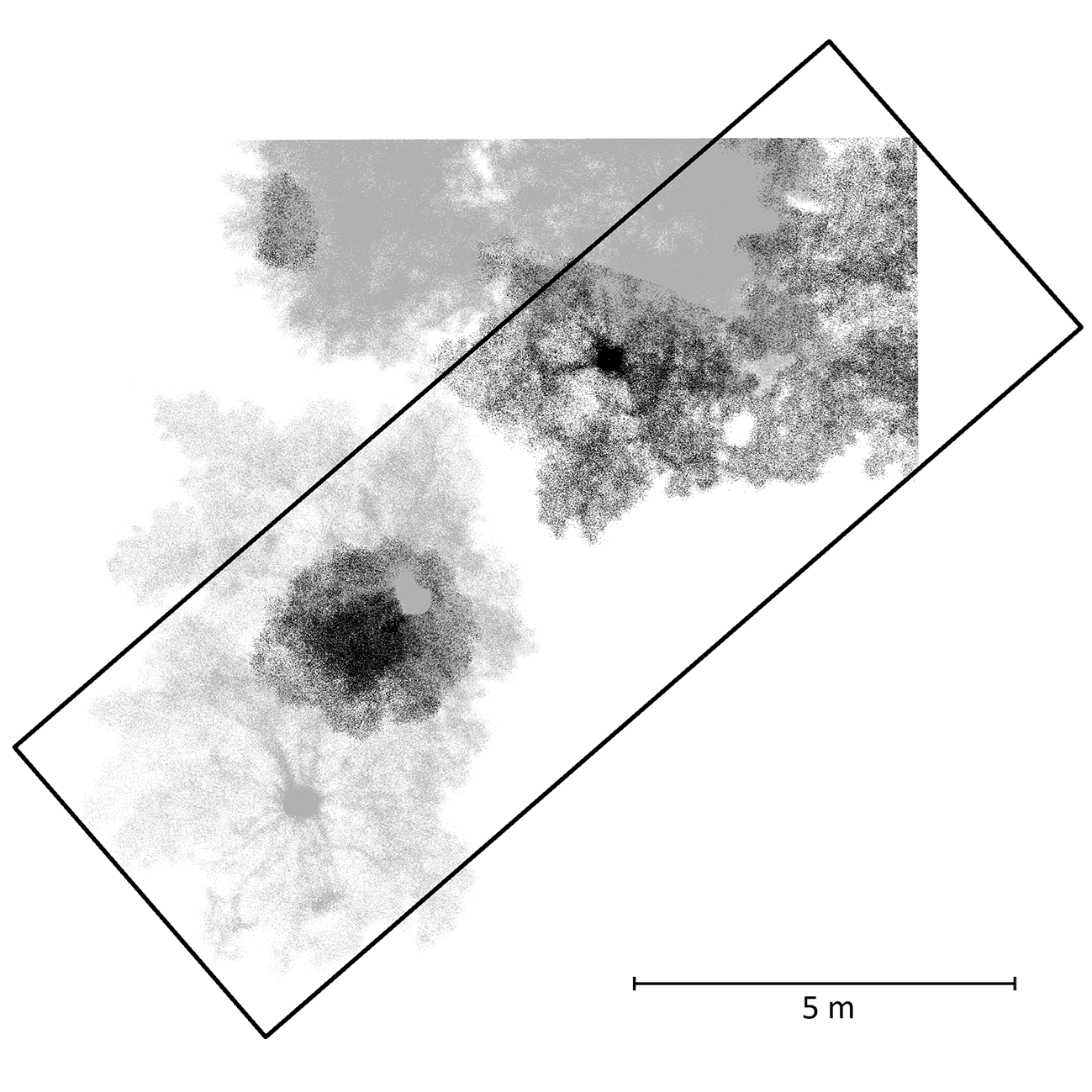}
    \end{subfigure}
    \begin{subfigure}[t]{0.4\linewidth}
    \centering
    \includegraphics[trim={290px 0 0 0}, clip, width=\linewidth]{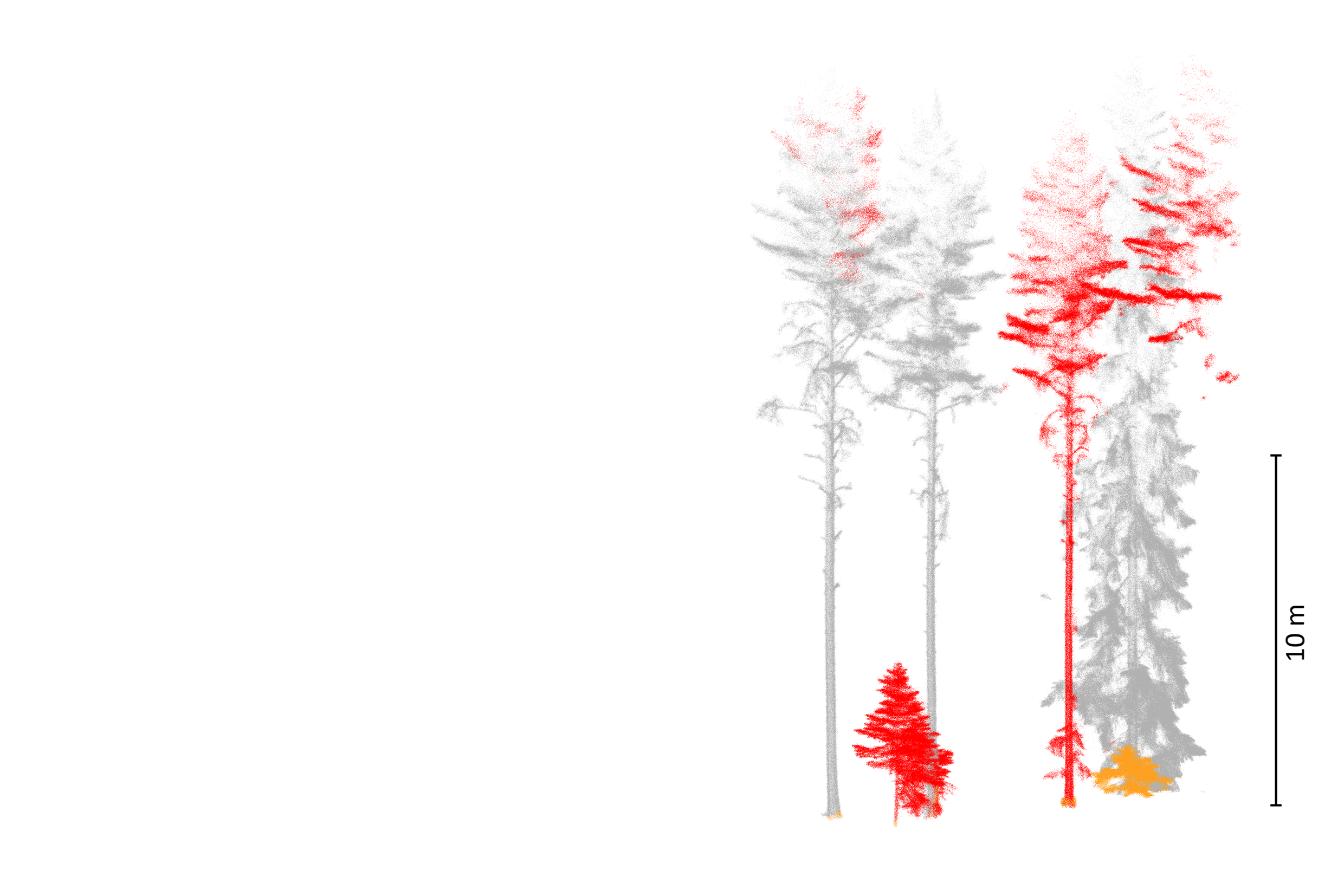}
    \end{subfigure}
    \caption{NIBIO MLS, test, plot 73}
    \label{fig:errors-nibio-mls}
    \end{subfigure}
    %
    %
    %
    \begin{subfigure}[t]{\linewidth}
    \begin{subfigure}[t]{0.25\linewidth}
    \centering
    \includegraphics[width=\linewidth]{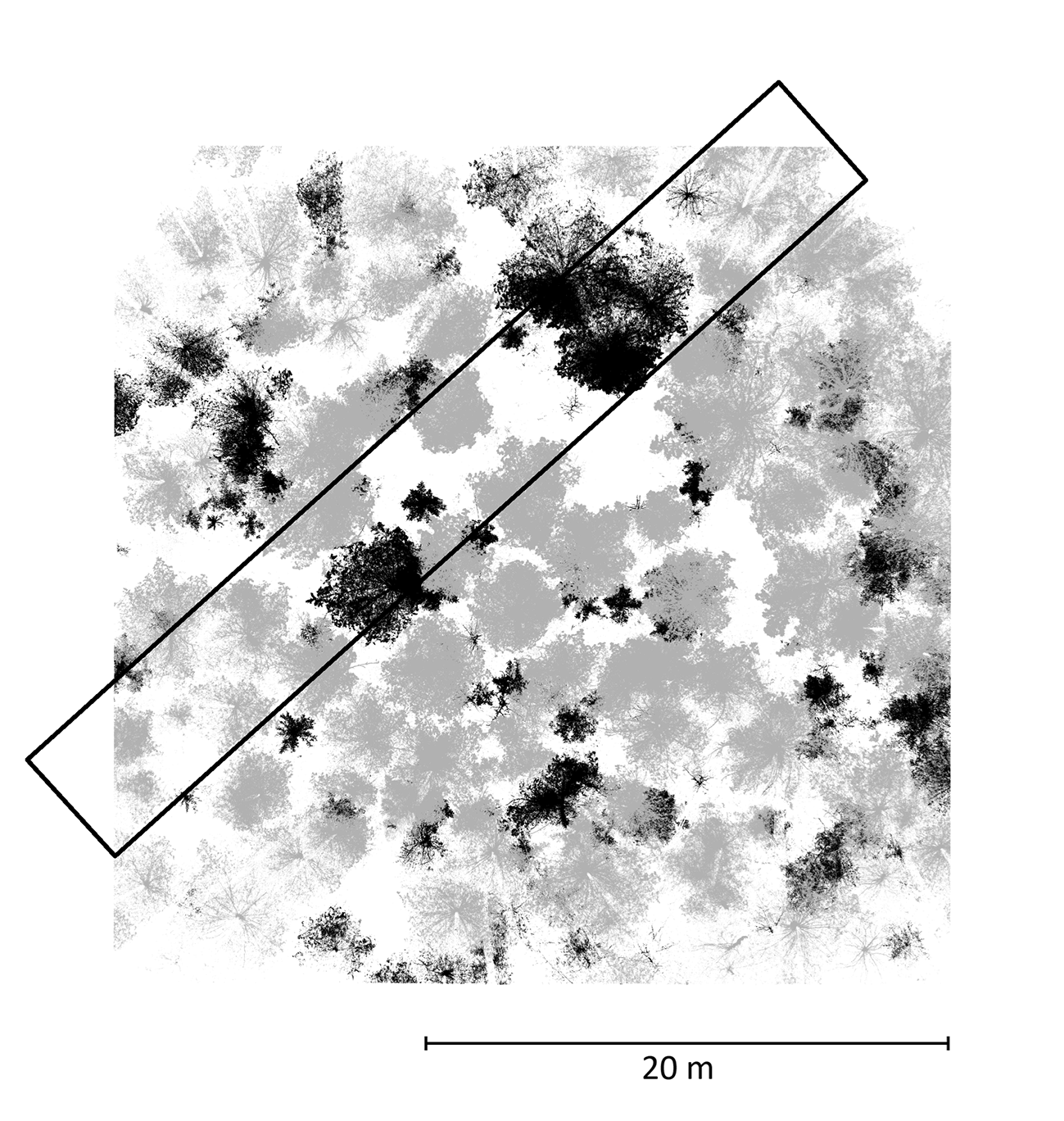}
    \end{subfigure}
    \begin{subfigure}[t]{0.5\linewidth}
    \centering
    \includegraphics[trim={0 30px 0 50px}, clip, width=\linewidth]{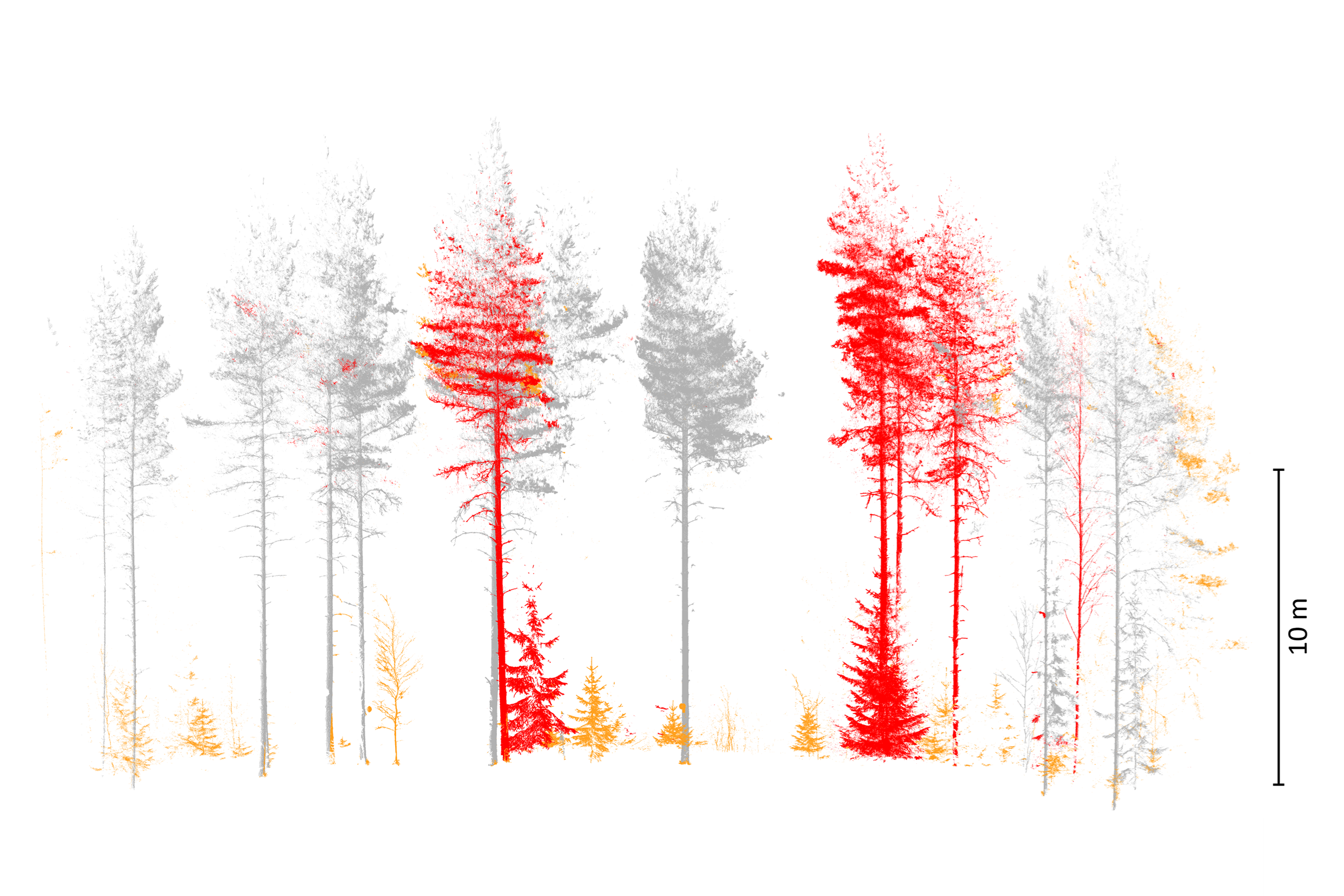}
    \end{subfigure}
    \caption{ForestSemantic, plot 3}
    \label{fig:errors-forest-semantic}    \end{subfigure}
    %
    %
    %
    \begin{subfigure}[t]{\linewidth}
    \begin{subfigure}[t]{0.29\linewidth}
    \centering
    \includegraphics[width=\linewidth]{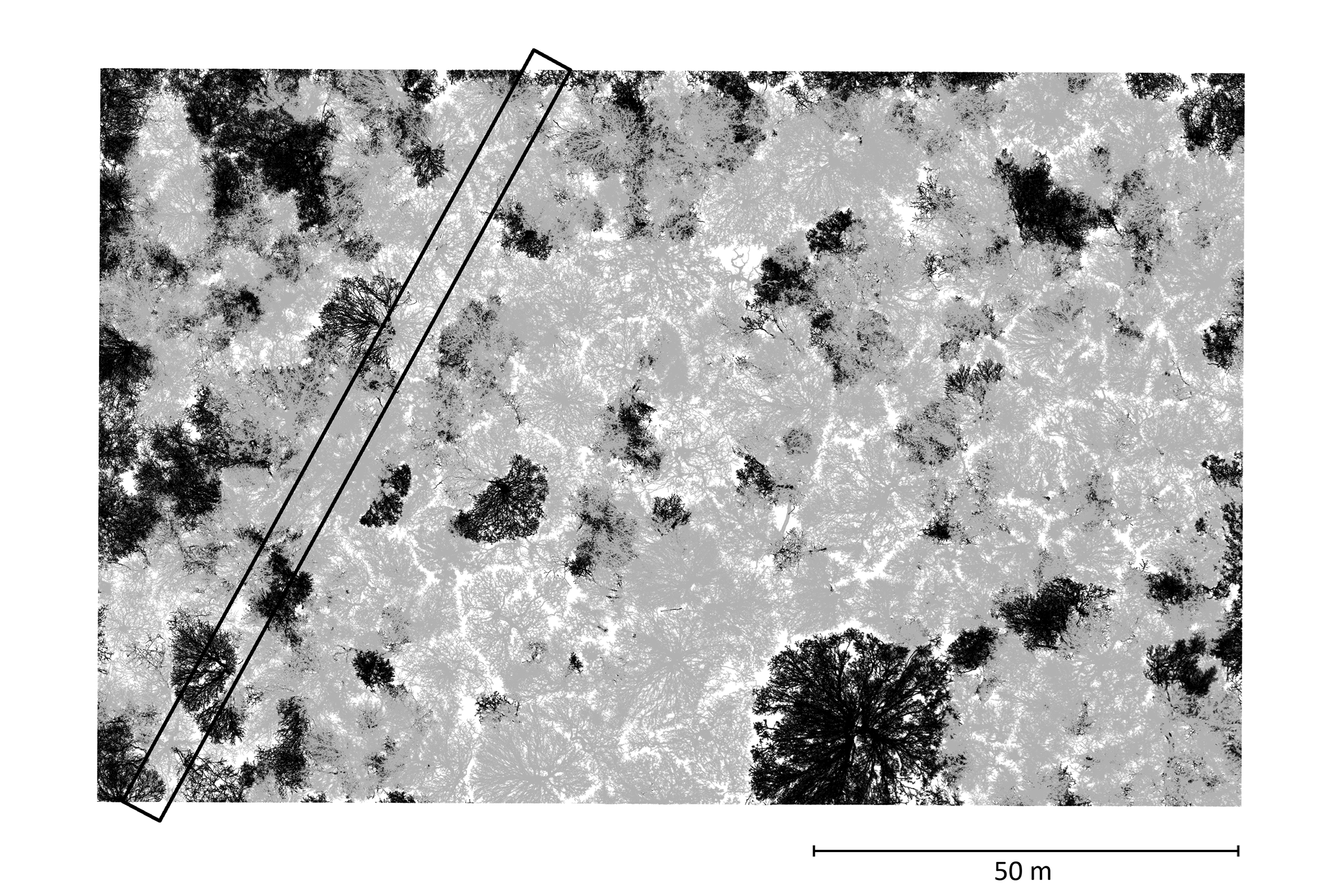}
    \end{subfigure}
    \begin{subfigure}[t]{0.69\linewidth}
    \centering
    \includegraphics[trim={0 50px 0 90px}, clip, width=\linewidth]{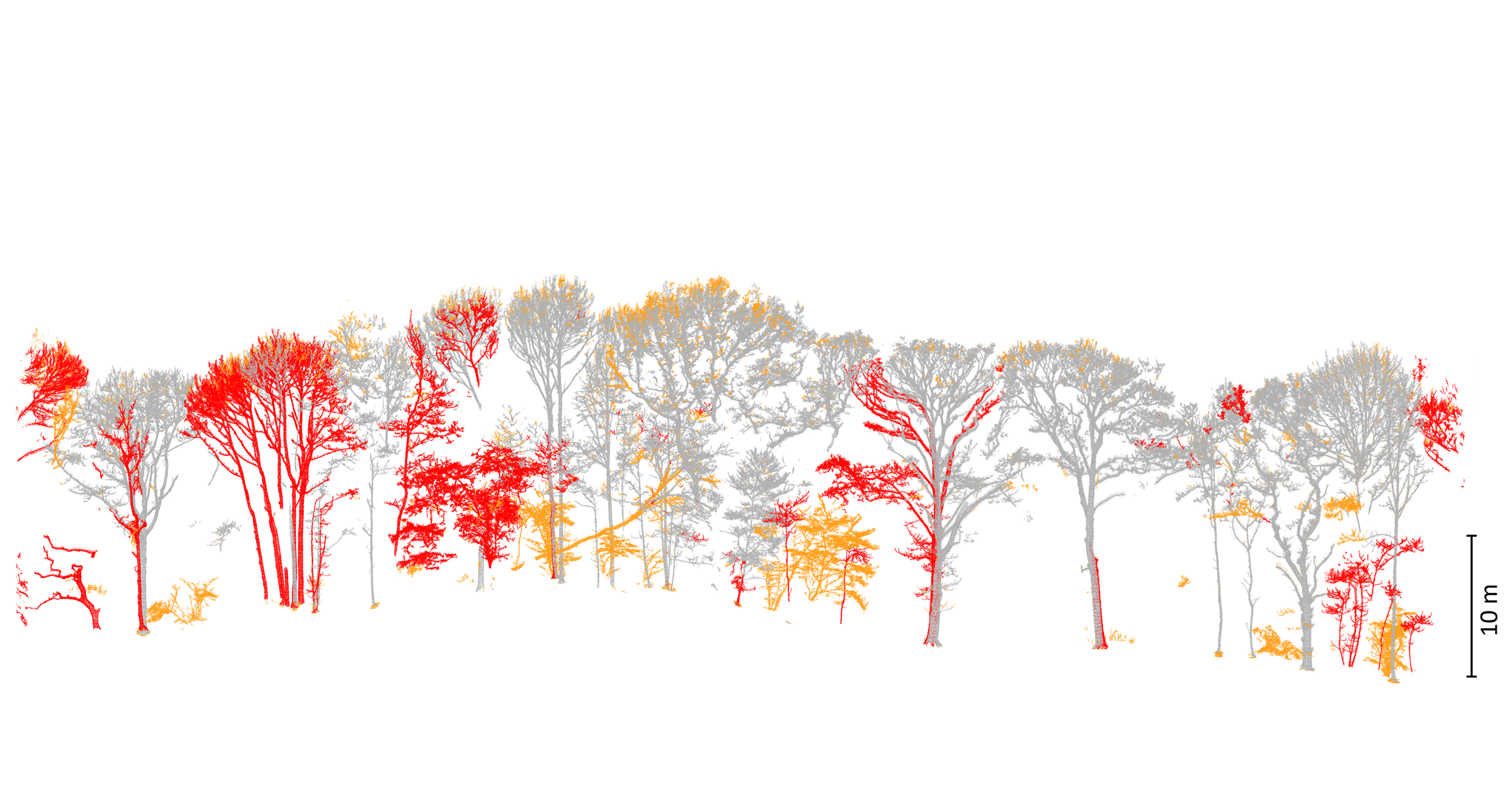}
    \end{subfigure}
    \caption{Wytham Woods}
    \label{fig:errors-wytham-woods}
    \end{subfigure}
    \caption{Errors of our algorithm on exemplary 3D~point clouds from the evaluation datasets. In the left column of each subfigure, an overview of the corresponding 3D point cloud is shown. In this overview, instance detection errors are visualized: correctly detected instances are shown in gray, while false positives and false negatives are shown in black. The right column of each subfigure provides a detailed view of a selected subsection of the respective 3D~point cloud. In the detailed view, instance segmentation errors are illustrated at the point level, including segmentation errors for instances that are true positives according to the instance detection metrics. Points that represent both omission and commission errors are shown in red, while those associated with only one of the two error types are shown in orange.}
    \label{fig:segmentation-errors}
\end{figure*}

%% file: fig_computational_resources_detailed.tex
\begin{figure}[ht]
\centering
\begin{scriptsize}
\begin{subfigure}[c]{\linewidth}
\vspace{0.18cm}
\centering
\includegraphics{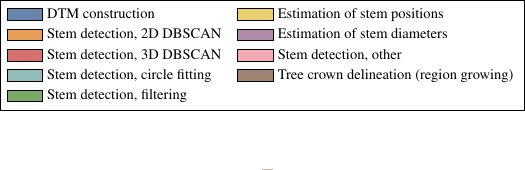}
\end{subfigure}
\begin{subfigure}[c]{\linewidth}
\vspace{-0.6cm}
\includegraphics{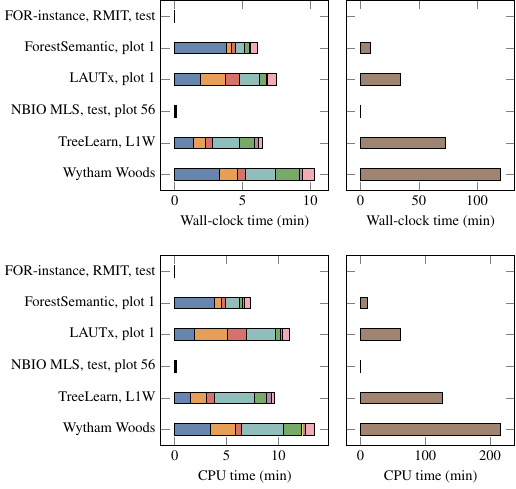}
\end{subfigure}
\end{scriptsize}
\caption{Runtime of the individual steps of our algorithm for exemplary 3D~point clouds. The runtime of the region growing step is shown in separate charts (right) since its runtime is considerably larger than that of the other steps. For the 3D~point cloud from the FOR-instance dataset, the runtime of the ULS preset is shown. For all other 3D~point clouds, the runtime of the TLS preset is shown.}
\label{fig:runtime-detailed}
\end{figure}

%% file: fig_computational_resources.tex
\begin{figure*}
\centering
\begin{scriptsize}
\begin{subfigure}[c]{\linewidth}
\centering
\includegraphics{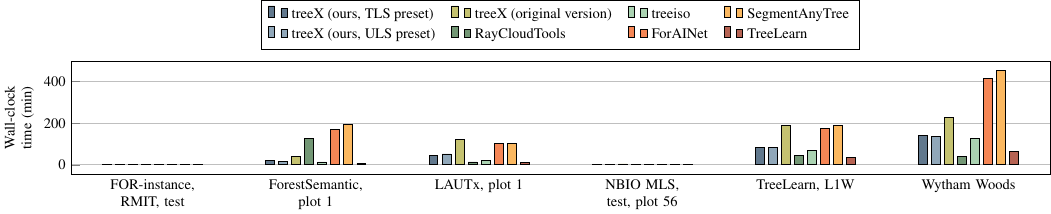}
\vspace{-0.2cm}
\caption{Wall-clock time}
\label{fig:wallclock-time}
\vspace{0.2cm}
\end{subfigure}
\begin{subfigure}[c]{\linewidth}
\centering
\includegraphics{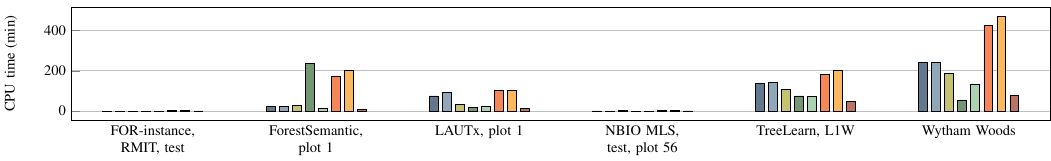}
\vspace{-0.2cm}
\caption{CPU time}
\label{fig:cpu-time}
\vspace{0.2cm}
\end{subfigure}
\begin{subfigure}[c]{\linewidth}
\centering
\includegraphics{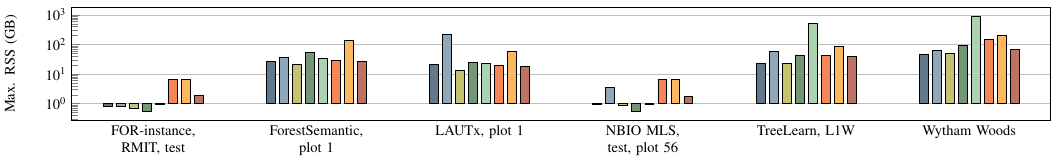}
\vspace{-0.2cm}
\caption{Maximum resident set size}
\label{fig:rss}
\vspace{0.2cm}
\end{subfigure}
\caption{Comparison of the computational resources used by our algorithm with existing open-source tree instance segmentation methods. Note the logarithmic scale in the bottom chart (max. RSS = maximum resident set size).}
\label{fig:computational-resources}
\end{scriptsize}
\end{figure*}

%% file: fig_ablation_study_tls.tex
\begin{figure*}
\centering
\begin{scriptsize}
\begin{subfigure}[c]{\linewidth}
\centering
\includegraphics{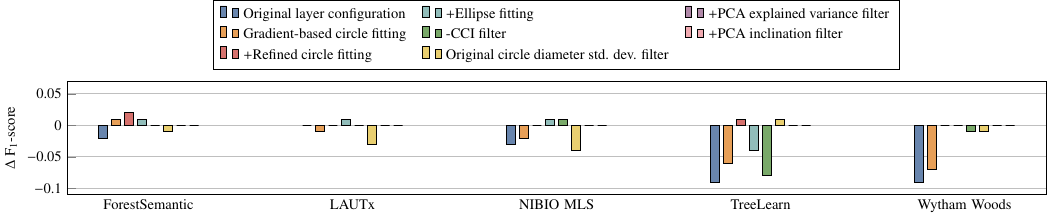}
\vspace{-0.1cm}
\caption{Difference in F\(_1\)-score}
\label{fig:ablation-study-tls-f1-score}
\vspace{0.1cm}
\end{subfigure}
\begin{subfigure}[c]{\linewidth}
\centering
\includegraphics{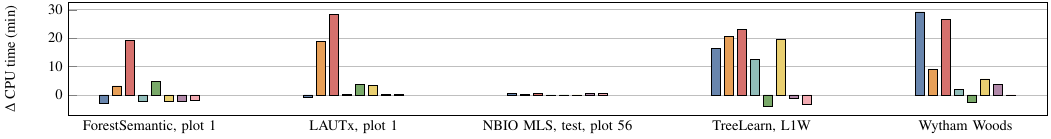}
\vspace{-0.1cm}
\caption{Difference in CPU time}
\label{fig:ablation-study-tls-cpu-time}
\vspace{0.1cm}
\end{subfigure}
\begin{subfigure}[c]{\linewidth}
\centering
\includegraphics{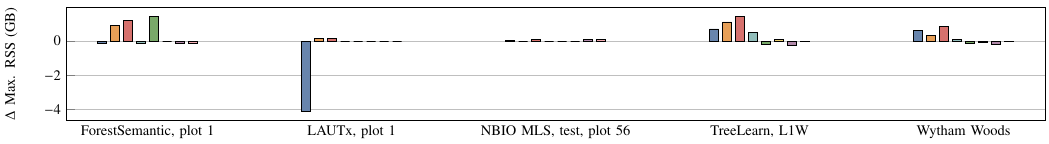}
\vspace{-0.1cm}
\caption{Difference in maximum resident set size}
\label{fig:ablation-study-tls-rss}
\vspace{0.1cm}
\end{subfigure}
\caption{Results of the ablation study with regard to the TLS preset of our algorithm. The charts display the changes in the metrics compared to the TLS preset. A zero value corresponds to the performance of the TLS preset. A bar in the positive direction indicates that the corresponding value increased in the ablation experiment compared to the TLS preset, while a bar in the negative direction indicates a decrease compared to the TLS preset. In the legend, “+” means that the respective component was added compared to the TLS preset and “-” that it was removed. Note that the F$_1$-score is aggregated over the entire datasets, while the CPU time and maximum resident set size are shown for exemplary point cloud files (max. RSS = maximum resident set size, std. dev. = standard deviation).}
\label{fig:ablation-study-tls}
\end{scriptsize}
\end{figure*}

%% file: fig_ablation_study_uls.tex
\begin{figure}
\centering
\vspace{0.18cm}
\begin{scriptsize}
\begin{subfigure}[c]{\linewidth}
\centering
\includegraphics{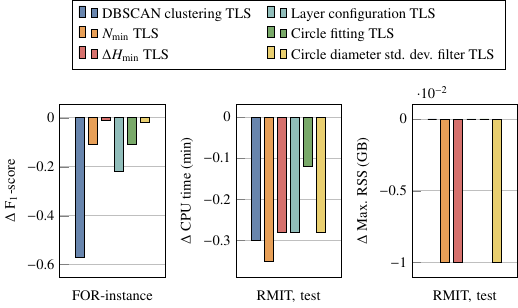}
\end{subfigure}
\caption{Results of the ablation study with regard to the ULS preset of our algorithm. The charts display the changes in the metrics compared to the ULS preset. A zero value corresponds to the performance of the ULS preset. A bar in the positive direction indicates that the corresponding value increased in the ablation experiment compared to the ULS preset, while a bar in the negative direction indicates a decrease compared to the ULS preset. In the legend, "TLS" means that the respective settings from the TLS preset were used. Note that the F$_1$-score is aggregated over the entire FOR-instance dataset, while the CPU time and maximum resident set size (RSS) are shown for an exemplary point cloud file, RMIT test, from the FOR-instance dataset.}
\label{fig:ablation-study-uls}
\end{scriptsize}
\end{figure}

%% file: sec_6_discussion.tex
Our evaluation shows that our revised version of the treeX algorithm substantially improves accuracy over the original version while also reducing runtime in terms of wall-clock time (mean instance detection F\(_1\)-score across all evaluation datasets: \num{0.68} for our version, \num{0.4} for the original version; wall-clock time for Wytham Woods dataset: \num{140}\,min for our version, \num{229}\,min for the original version).
As demonstrated by the ablation studies, these gains are primarily due to refinements of the parameter settings of the stem detection stage and a more efficient circle fitting procedure.
The core principle of the algorithm remains unchanged, with clustering-based stem detection forming the starting point for tree instance segmentation.
Due to this design, our method is best suited for dense ground-based \ac{TLS} and \ac{PLS} scans, where stems are typically well represented.
In contrast, the sparser representation of below-canopy areas in \ac{ULS} point clouds increases the likelihood of stem detection errors and thus limits the broader applicability of our approach to \ac{ULS} data.
However, the evaluation of the \ac{ULS} preset demonstrates that our algorithm can also perform well on \ac{ULS} data, provided that the flight pattern and canopy structure enable the acquisition of point clouds with sufficient point density in the stem layer.

Overall, the accuracy of our  algorithm is competitive with existing open-source tree instance segmentation methods.
A comparison with treeiso and RayCloudTools, two other unsupervised methods, shows that our algorithm offers improved generalizability across datasets (mean instance detection F\(_1\)-score across all evaluation datasets: \num{0.68}~for our algorithm, \num{0.5}~for RayCloudTools, \num{0.47}~for treeiso). Although RayCloudTools achieves high accuracy and good computational performance for several datasets, it produces more errors on the ForestSemantic dataset. While most existing open-source unsupervised tree instance segmentation algorithms, including RayCloudTools and treeiso, are graph-based, our algorithm uses a conceptually different approach that combines density-based clustering with region growing. In addition, our algorithm is more thoroughly documented and tested than most existing open-source alternatives (Table~\ref{tab:related-work}), making it a valuable complement to the landscape of unsupervised algorithmic tree instance segmentation methods.

Compared to recent deep learning approaches, the overall segmentation accuracy of our algorithm is on a similar level (mean instance detection F\(_1\)-score across all evaluation datasets: \num{0.68} for our algorithm, \num{0.67} for ForAINet, \num{0.63} for Segment\-Any\-Tree, \num{0.6} for TreeLearn). However, deep learning methods generally outperform our algorithm and the other tested algorithmic methods when applied to sparse point cloud data, such as the FOR-instance dataset. A key advantage of deep learning approaches is their ability to aggregate information over larger receptive fields and to analyze spatial patterns across multiple scales~\citep{henrich-2024}. This reduces their dependence on local features, such as stem sections within a specific height range, which is a common limitation of algorithmic tree instance segmentation methods~\citep{liang-2018}. In the long term, we consider deep learning to be a promising direction for tree instance segmentation, particularly due to its potential for better generalizability across diverse forest stand structures and varying sensor types~\citep{wielgosz-2024}. This includes challenging conditions such as dense, multi-layered forests and sparse point clouds with strong occlusions. However, our evaluation shows that current deep learning models do not fully achieve this level of generalization. Their performance is closely tied to the composition of their training data. For example, ForAINet, which was trained on the FOR-instance dataset, and SegmentAnyTree, which was trained on both the FOR-instance and NIBIO MLS datasets, outperform other methods on these specific datasets. Conversely, TreeLearn, which was pre-trained on a large corpus of deciduous stands in addition to the FOR-instance dataset, outperforms other deep learning methods on deciduous-dominated datasets such as LAUTx, TreeLearn, and Wytham Woods. Surprisingly, however, TreeLearn performs poorly on the NIBIO MLS dataset. This may be because we used the \enquote{small} TreeLearn checkpoint in our evaluation to ensure a fair comparison, and the TreeLearn checkpoints trained on larger datasets may yield improved performance~\citep{henrich-2024}.
Nevertheless, this demonstrates that the out-of-the-box generalizability of deep learning models is difficult to anticipate. Thus, a diverse, standardized, and fully described evaluation dataset, categorized by stand characteristics and sensor and platform types, is needed to benchmark tree instance segmentation approaches~\citep{cherlet-2024}. This will enable a comprehensive evaluation across diverse stand conditions and assess the ability to generalize.

Considering the aforementioned limitations, our algorithm could contribute to improving the robustness of deep learning approaches in several ways. First, it can serve as a tool for automatic or semi-automatic labeling of training data, supporting the creation of more diverse and balanced training datasets. The advantage of using algorithmic methods to generate training data, rather than deep learning models, is that they exhibit different error patterns that are independent of current deep learning models, reducing the risk of reinforcing existing errors.
Second, our algorithm could be integrated into hybrid methods that combine algorithmic and deep learning approaches. For example, the stem detection stage of our algorithm could incorporate a deep-learning-based semantic segmentation, similar to TLS2trees~\citep{wilkes-2023}.
Furthermore, deep-learning-based semantic segmentation could be used to prioritize wood points over leaf points in the region growing process of our algorithm.
Alternatively, the stem positions detected by our algorithm could serve as instance queries in transformer-based instance segmentation models, such as TreeisoNet~\citep{xi-2025} or ForestFormer3D~\citep{xiang-2025}.

Complementary to such integrations with deep learning, our method can also serve as a resource-efficient alternative in scenarios where the characteristics of the input data are known and align with the strengths of our algorithm. This is particularly the case for \ac{TLS} and \ac{PLS} point clouds with good stem visibility. Compared to deep learning approaches, our algorithm offers higher computational efficiency (notably, no GPU is required) and higher interpretability. Its rule-based design makes its behavior on new datasets more predictable. This allows for assessing its suitability for specific application scenarios and adjusting its parameters if necessary.

To support broader adoption, we provide a robust and well-tested Python implementation of our algorithm as part of the pointtree package. This facilitates integration into larger processing pipelines. In the future, this implementation could be further enhanced by adding tile-based, out-of-core processing to enable the handling of massive point cloud datasets without being constrained by main memory limitations.

%% file: sec_7_conclusions.tex
In this paper, we presented a revised version of the treeX algorithm, which was originally introduced by \citet{tockner-2022} and \citet{gollob-2020}, for unsupervised tree instance segmentation in dense forest point clouds. The algorithm combines a stem detection procedure based on density-based point clustering and a series of filtering criteria with a region growing approach for crown delineation.
Our revision substantially improved the accuracy of the original method, while also reducing its runtime through better parallelization.
Our main modifications comprise a reparameterization of the stem detection stage, the implementation of a more efficient circle fitting procedure, and the introduction of different parameter presets for ground-based and UAV-borne data. The effectiveness of these modifications was demonstrated in ablation studies.
The design of our method is particularly effective when applied to dense \ac{TLS} and \ac{PLS} point clouds of stands with good stem visibility, i.e., stands with few branches near the ground and a loose to moderate understory. For such data, it performs competitively against other publicly available tree instance segmentation methods, including recent deep learning approaches.

We see two main areas of application for our method: First, it offers a computationally efficient and interpretable alternative to deep learning approaches for input data with the aforementioned characteristics. Second, it can serve as a tool for the automatic or semi-automatic generation of labeled training data, supporting the development of more generalizable deep learning methods. Future work could also explore combinations of our algorithm with deep learning approaches, for example, by integrating deep-learning-based semantic segmentation into our algorithm or using the detected stem positions as prompts for deep-learning-based instance segmentation models.

To support adoption and integration into larger processing pipelines, we provide a well-tested and well-documented implementation in an open-source Python package. In this way, our algorithm complements the current landscape of open-source tree instance segmentation methods for dense forest point clouds.

%% file: sec_8_ai_disclosure.tex
During the preparation of this work, the authors used Open\-AI ChatGPT, DeepL, Writefull, and Grammarly to improve readability and language. After using these tools, the authors reviewed and edited the content as needed and take full responsibility for the content of the publication.

%% file: sec_9_authorship.tex
\textbf{Josafat-Mattias Burmeister}: Conceptualization; Methodology; Software; Formal analysis; Investigation; Data Curation; Writing - Original Draft; Visualization. \textbf{Andreas Tockner}: Conceptualization; Methodology; Data Curation; Writing - Review \& Editing; Visualization. \textbf{Stefan Reder}: Conceptualization; Methodology; Data Curation; Writing - Review \& Editing. \textbf{Markus Engel}: Conceptualization; Writing - Original Draft; Supervision. \textbf{Rico Richter}: Conceptualization; Writing - Review \& Editing; Supervision; Project administration; Funding acquisition. \textbf{Jan-Peter Mund}: Writing - Review \& Editing; Supervision; Project administration; Funding acquisition. \textbf{Jürgen Döllner}: Writing - Review \& Editing; Supervision; Project administration; Funding acquisition.

%% file: sec_10_declaration_of_competing_interest.tex
The authors declare that they have no known competing financial interests or personal relationships that could have appeared to influence the work reported in this paper.

%% file: sec_11_acknowledgements.tex
This work was partially funded by the Federal Ministry of Research, Technology and Space, Germany through grant 033L305 (\enquote{TreeDigitalTwins}) and grant 01IS22062 (AI research group \enquote{FFS-AI}).

We would like to thank Sebastian Mitte for providing a prototypical reimplementation of the original treeX algorithm in Python, which was used as a reference in the development of our implementation of the treeX algorithm.

%% file: sec_12_data_availability.tex
\sloppy
The code of the presented tree instance segmentation method is published on GitHub: \url{https://github.com/ai4trees/pointtree}.
The code of the reference tree instance segmentation methods is publicly available: original version of treeX: \url{https://github.com/anditockner/treeX}; treeiso: \url{https://github.com/truebelief/artemis_treeiso}; RayCloudTools: \url{https://github.com/csiro-robotics/raycloudtools}; ForAINet: \url{https://github.com/prs-eth/ForAINet}; SegmentAnyTree: \url{https://github.com/SmartForest-no/SegmentAnyTree}; TreeLearn: \url{https://github.com/ecker-lab/TreeLearn}.

The data used in this study are publicly available: FOR-instance dataset: \url{https://doi.org/10.5281/zenodo.8287791}; ForestSemantic dataset: \url{https://doi.org/10.5281/zenodo.13285639}; LAUTx dataset: \url{https://doi.org/10.5281/zenodo.6560111}; NIBIO MLS dataset: \url{https://doi.org/10.5281/zenodo.12754725}; TreeLearn dataset: \url{https://doi.org/10.25625/VPMPID}; Wytham Woods dataset: \url{https://doi.org/10.25625/QUTUWU}.

%% file: sec_13_appendix.tex
\section{Circle Fitting Methods}
\label{sec:circle-fitting-details}

For fitting circles to sets of 2D~points, as done in the stem detection stage of our algorithm, we evaluate two outlier-robust circle detection methods: the gradient-based method proposed by \citet{garlipp-2006} and a method based on \ac{RANSAC}~\citep{fischler-1981}. The former corresponds to the approach used in the original version of the treeX algorithm, whereas the latter was explored as an alternative in our study.
Although both methods are capable of detecting multiple circles in noisy point sets, we use them for circle fitting, retaining only the circle with the highest goodness-of-fit.
Both methods use an objective function based on that introduced by \citet{garlipp-2006} to measure the goodness-of-fit of the detected circles:
\begin{equation}
\label{eqn:circle-fitting-objective}
\mathcal{S}\left(a, b, r\right) = \sum_{i=1}^N \frac{1}{s} \rho
\left(
\frac{\left\|\begin{bmatrix}x_i, y_i \end{bmatrix}^T - \begin{bmatrix} a, b \end{bmatrix}^T\right\| - r}{s}
\right)
\end{equation}
Here, \([a, b]\) are the xy-coordinates of the circle center and \(r\) is the circle radius. \(N\) is the number of input points and \([x_i,y_i]\) are the xy-coordinates of the i-th point. The function \(\rho\) is a kernel function, implemented as a Gaussian in our case. \(s\) is the kernel bandwidth, which is a user-defined parameter of our algorithm (Table~\ref{tab:stem-detection-parameters}).
Using such a kernel-based objective function reduces the influence of points with large residuals, making it more robust to outliers. In contrast to the original objective function proposed by \citet{garlipp-2006}, we use a sum rather than a mean to aggregate the per-point scores, thereby giving preference to circles supported by a larger number of points.

\subsection{Gradient-Based Method}

The gradient-based method proposed by \citet{garlipp-2006} begins with a set of initial circle parameters and iteratively fits them to the input points using a gradient-based optimization method, either Newton's method or gradient descent.
The loss function minimized during this process is based on the objective function in Eq.~\ref{eqn:circle-fitting-objective}:
\begin{equation}
\label{eqn:circle-fitting-loss}
\mathcal{L}\left(a, b, r\right) = - \frac{1}{N} \mathcal{S}\left(a, b, r\right)
\end{equation}
Since the loss function is minimized, the negative of the objective function is used, and the sum is replaced by a mean to improve numerical stability.

To enable the detection of circles at various positions and scales, the optimization is repeated for multiple initializations of the circle parameters.
Nine different initializations of the circle center position are tested, arranged as a regular \(3 \times 3\) grid covering the axis-aligned bounding box of the input points.
For each center position, three initial values for the radius are tested: \(\diameter_{\min}\), \(\diameter_{\max}\), and \(\diameter_{\min} + 0.5 \cdot (\diameter_{\max} - \diameter_{\min})\) where \(\diameter_{\min}\) and \(\diameter_{\max}\) are user-defined parameters of our algorithm (Table~\ref{tab:stem-detection-parameters}).

To filter out candidate circles for which the optimization is not converging, constraints are imposed on the parameter values during optimization.
Specifically, the circle diameter must remain within the bounds \(\diameter_{\min}\) and \(\diameter_{\max}\).
The circle center must remain within the axis-aligned bounding box of the input points, expanded by a small margin.
If the circle parameters leave these bounds during optimization, the optimization is terminated, and the candidate circle is discarded.
Otherwise, optimization continues until a maximum number of \num{1000} iterations is reached or the magnitude of the parameter update falls below a threshold of \(10^{-5}\).
All circles for which the optimization converges and the goodness-of-fit is above a user-defined threshold \(\mathcal{S}_{\min}\) are returned (Table~\ref{tab:stem-detection-parameters}).

\subsection{RANSAC Method}

\ac{RANSAC}~\citep{fischler-1981} is a general-purpose, iterative model fitting approach in which a model is fitted to a randomly sampled subset of data points in each iteration.
This subset, referred to as the hypothetical inliers, is used to generate a candidate model.
For each candidate model, a consensus set is determined, consisting of all data points whose deviation from the model falls within a predefined error tolerance.
If the size of the consensus set exceeds a predefined threshold, the model is considered a valid candidate and is re-estimated using all points in the consensus set.
We adopt and extend this approach to detect circles in 2D~point sets.
To this end, we perform the following five steps in each RANSAC iteration:

\begin{enumerate}
    \item A set of hypothetical inliers is randomly sampled from the set of input points. We set the size of the sampled subset to three, which is the minimum number of points required to uniquely define a circle. Using the smallest possible sample size increases the variability of circle candidates across iterations.
    \item A candidate circle is fitted to the hypothetical inliers using a least-squares method. If the parameters of the fitted circle are outside a user-specified range, the candidate is immediately rejected, and the algorithm proceeds to the next iteration. Specifically, the circle diameter must lie within the range \(\diameter_{\min}\) and \(\diameter_{\max}\) (Table~\ref{tab:stem-detection-parameters}), and the circle center is constrained to lie within the axis-aligned bounding box of the input points, expanded by a small margin.
    \item If the candidate circle passes the parameter constraints, all points from the original set of input points are tested against the circle. Points whose distance to the circle's outline is below an error threshold are included in the consensus set. The error threshold is set to the same value as the kernel bandwidth \(s\) of the objective function in our implementation. 
    \item The circle is re-fitted using all points in the consensus set to improve the goodness-of-fit.
    \item The goodness-of-fit of the re-fitted circle is evaluated using the objective function in Eq.~\ref{eqn:circle-fitting-objective}. All circles with a goodness-of-fit above a user-defined threshold \(\mathcal{S}_{\min}\) (Table~\ref{tab:stem-detection-parameters}) and a consensus set containing at least three points are accepted as valid and returned.
\end{enumerate}

\subsection{Filtering of Detected Circles}

Although both circle detection methods are capable of detecting multiple circles in a given input point set, we are only interested in retrieving the circle with the highest goodness-of-fit.
To identify the best-fitting circle, we apply the following steps:
\begin{enumerate}
    \item Circles whose parameters are equal up to the fourth decimal place are deduplicated, retaining only one of the circles.
    \item Overlapping circles are filtered using non-maximum suppression. From a set of overlapping circles, only the circle with the highest goodness-of-fit according to Eq.~\ref{eqn:circle-fitting-objective} is retained.
    \item The \ac{CCI} \citep{krisanski-2020} is calculated for each remaining circle. Only circles with a \ac{CCI} above a user-defined threshold \(CCI_{\min}\) are retained (Table~\ref{tab:stem-detection-parameters}). If \(CCI_{\min}\) is set to \enquote{None}, this filtering step is skipped.
    The \ac{CCI} quantifies how well the outline of a fitted circle is covered by input points.
    It is calculated by dividing the circle into a set of angular regions.
    An angular region is marked as complete if it contains at least one point whose distance to the circle outline is equal to or less than a user-specified error threshold.
    The \ac{CCI} is then defined as the proportion of angular regions that are complete.
    We set the number of angular regions to \num{73} and the error threshold to the same value as the kernel bandwidth \(s\) of the objective function.
    \item From the remaining circles, the circle with the highest goodness-of-fit according to Eq.~\ref{eqn:circle-fitting-objective} is selected.
\end{enumerate}

To avoid circle fitting on very sparse inputs, we define a threshold \(N_{\min}^{circ}\), representing the minimum number of input points required for circle fitting (Table~\ref{tab:stem-detection-parameters}). If the number of input points falls below this threshold, or if the fitting process is not successful, no circle is returned. Otherwise, the circle selected through the aforementioned process is returned.

\subsection{Implementation Details}

The circle fitting methods was implemented in the Python package circle-detection~\citep{burmeister-2025a}.\footnote{Repository of the circle-detection package: \url{https://github.com/josafatburmeister/circle_detection}} To improve performance, large parts of the methods were implemented in C++ and integrated with the Python code via pybind11~\citep{jakob-2025}.\footnote{Repository of pybind11: \url{https://github.com/pybind/pybind11}}